\documentclass[lettersize,journal]{IEEEtran}
\usepackage{amsmath,amsfonts}
\usepackage{algorithmic}
\usepackage{algorithm}
\usepackage{array}
\usepackage[caption=false,font=normalsize,labelfont=sf,textfont=sf,labelfont=rm,textfont=rm]{subfig}
\usepackage{textcomp}
\usepackage{stfloats}
\usepackage{url}
\usepackage{verbatim}
\usepackage{graphicx}
\usepackage{cite}
\usepackage{booktabs}    
\usepackage{utfsym}
\usepackage{multirow}
\usepackage{makecell}
\usepackage{color}
\usepackage{mathptmx}
\def\BibTeX{{\rm B\kern-.05em{\sc i\kern-.025em b}\kern-.08em
    T\kern-.1667em\lower.7ex\hbox{E}\kern-.125emX}}
\usepackage{balance}
\begin{document}
\title{
	Multi-Level Heterogeneous Knowledge Transfer Network on Forward Scattering Center Model for Limited Samples SAR ATR}
\author{Chenxi Zhao, Daochang Wang, Wancong Li, Siqian Zhang, Siyuan He, Gangyao Kuang, ~\IEEEmembership{Senior Member,~IEEE}
\thanks{(Corresponding author: Siqian Zhang.)
	
Chenxi Zhao, Daochang Wang, Siqian Zhang and Gangyao Kuang from College of Electronic Science and Technology, National University of Defense Technology, Changsha, 410073, China.
Wancong Li and Siyuan He from School of Electronic Information, Wuhan University, Wuhan, 430072, China.
}}

\maketitle

\begin{abstract}
Simulated data-assisted SAR target recognition methods are the research hotspot currently, devoted to solving the problem of limited samples.
Existing works revolve around simulated images, but the large amount of irrelevant information embedded in the images, such as background, noise, etc., seriously affects the quality of the migrated information.
Our work explores a new simulated data to migrate purer and key target knowledge, i.e., forward scattering center model (FSCM) which models the actual local structure of the target with strong physical meaning and interpretability.
To achieve this purpose, multi-level heterogeneous knowledge transfer (MHKT) network is proposed, which fully migrates FSCM knowledge from the feature, distribution and category levels, respectively.
Specifically, we permit the more suitable feature representations for the heterogeneous data and separate non-informative knowledge by task-associated information selector (TAIS), to complete purer target feature migration. 
In the distribution alignment, the new metric function maximum discrimination divergence (MDD) in target generic knowledge transfer (TGKT) module perceives transferable knowledge efficiently while preserving discriminative structure about classes. 
Moreover, category relation knowledge transfer (CRKT) module leverages the category relation consistency constraint to break the dilemma of optimization bias towards simulation data due to imbalance between simulated and measured data.
Such stepwise knowledge selection and migration will ensure the integrity of the migrated FSCM knowledge.
Notably, extensive experiments on two new datasets formed by FSCM data and measured SAR images demonstrate the superior performance of our method.
\end{abstract}


\section{Introduction}
\subsection{Background}
Synthetic Aperture Radar (SAR) provides reliable imaging irrespective of weather and time of day, e.g., stable imaging under clouds and fog as well as changes in light.
With the gradually higher resolution of SAR images, SAR target recognition has caught wider attention \cite{9064495,10570499}.

During the past decade, deep learning methods \cite{7460942, 7780459, simonyan2014very, 7298594} have been successful in various fields and have become the dominant method in SAR automatic target recognition (ATR).
Normally, an excellent deep model demands extensive labelled data as support \cite{7926358}. 
However, given the challenges of acquiring and labeling SAR data, it is hard to satisfy the data demands of the model.
Therefore, SAR ATR under limited samples has received widespread focus.

At first, researchers hope to find a solution for SAR ATR under limited samples from the deep learning model structure \cite{10335748, 10614616} and inherent properties of the SAR data \cite{10137878}.
However, such methods of modifying the network structures fail to break the limitations imposed by network itself.
Thus, researchers have gradually turned to the study of SAR inherent properties, which describe vital information about the target in the network.
Specifically, given that scattering center (SC) models SAR targets effectively, methods integrating SC features are gradually emerging as the mainstream \cite{10700786}.

Recently, the use of simulated data to assist SAR ATR under limited sample situations has become a hot research topic \cite{lewis2019sar,ZHANG2024103707}. 
One of the pioneering works on simulation data-assisted SAR ATR is \cite{7968358}.
It combines pre-training techniques and simulated images, which provides better initial model parameters for the measured images.
The feasibility of simulated image-assisted SAR ATR is demonstrated for the first time.
Additionally, domain adaptation (DA) \cite{WANG2018135} is a desirable solution to model the knowledge migration procedure from simulated to measured data.
Generally, we hypothesise that the source and target domains (SD and TD) correspond to simulated and measured data, respectively.
The purpose of DA is to mine different but related domain knowledge from the SD to improve classification performance and minimize the dependence on TD labelled data.
DA based methods have been widely employed in SAR ATR.
Most effective methods rely on the use of maximum mean discrepancy (MMD) \cite{NIPS2006_e9fb2eda,LI2023272}, as well as variants of MMD methods \cite{10217035}, or adversarial learning methods \cite{SHI20241}, etc.
In brief, the crucial concept of these methods is to learn domain-invariant features by minimizing divergence or fuzzy domain discriminators.  

\subsection{Motivation}
\begin{figure}[!t]
	\centering \includegraphics[width=3.4in]{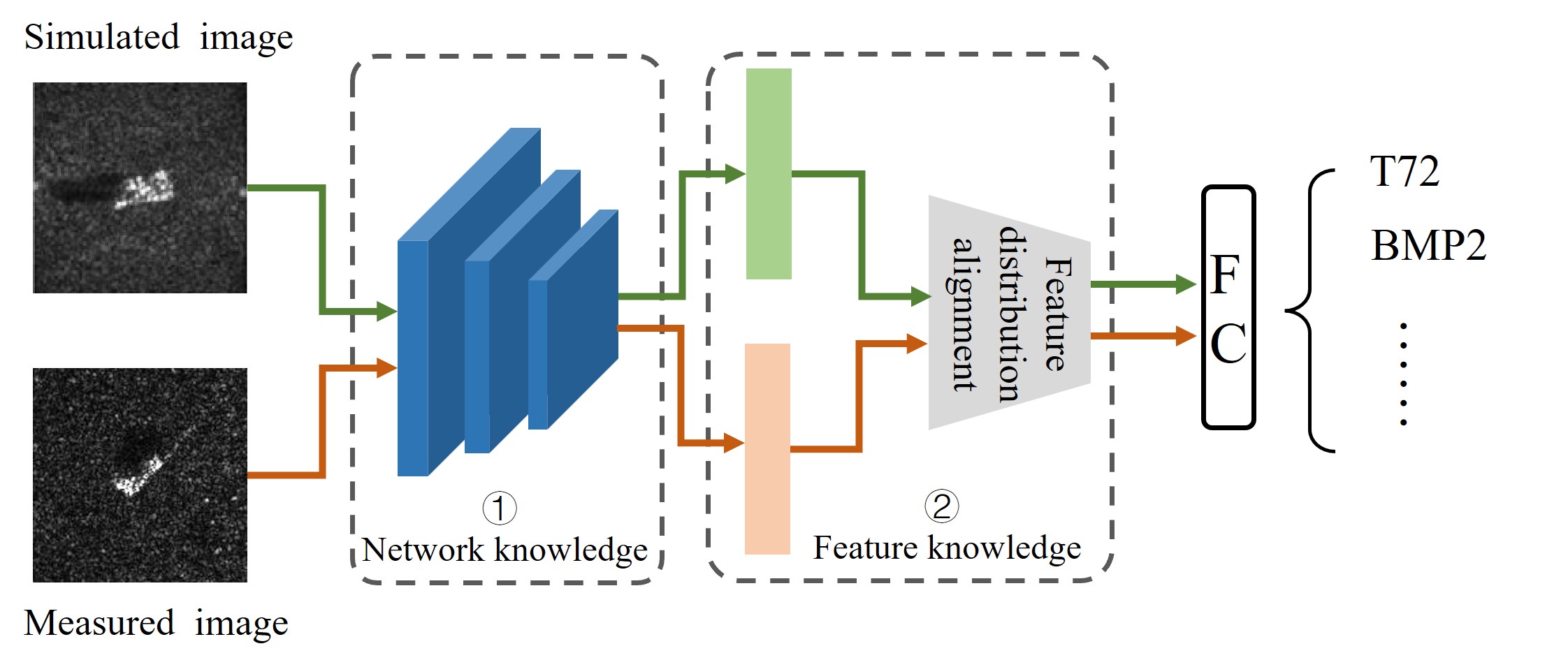}%
	\caption{Illustration of the knowledge migration process from simulated images.}
	\label{figI1}
\end{figure}

Existing methods have mainly studied the migration of knowledge from simulated images, and the transfer procedure of knowledge is illustrated in Fig. \ref{figI1}.
This migration framework suffers from the following problems:
\begin{itemize}
	\item [i)] The current methods focus primarily on migrating simulated image knowledge.
	However, simulated images contain useless information such as background other than the target, which generate the serious risk of migrating unwanted information.
	\item [ii)] These methods have a crude knowledge migration process, forcing all domains to extract the features by exploiting the same network, as shown in Fig. \ref{figI1}, which works against maximizing the unique strengths of the domain.
\end{itemize}

These issues inspire us to ponder the following:
\begin{itemize}
	\item [1)] On the usage of simulated SCs. 
	Current methods demonstrate the prominent role of SC features in SAR ATR.
	SC features provide information about the local structure of the target, radar parameters, etc.
	They describe the essential properties of the target and provide greater discrimination.
	Moreover, as the research continues, scholars have developed studies on the simulated SCs, i.e. FSCM \cite{10364767,9091832}, which offer more explicit physical implications than SC features extracted from measured images.
	Compared to simulated SAR images, FSCM describes purer and vital target knowledge, which naturally avoids the negative impacts of background, noise and other irrelevant factors in the image.
	Therefore, it is reasonable to believe that FSCM knowledge will provide more stable, discriminative knowledge for the SAR ATR task.
	\item [2)] On the migration of heterogeneous features.
	The current studies focus on homogeneous domain transfer learning methods between images, as shown in Fig. \ref{figI1}.
	Such methods require that the SD and TD have the same feature semantics and dimensions.
	However, as previously mentioned, the SC features describe the electromagnetic scattering properties of the SAR target and models its local scattering phenomena.
	SAR images contain the target's geometry, texture, etc.
	There exists the most appropriate expression of each feature.
	Such heterogeneity between FSCM and measured SAR images invalidates such frameworks.
	Thus, exploring a knowledge migration method to fully transfer the heterogeneous FSCM knowledge to the measured SAR images is imperative.
	\item [3)] On the multi-level knowledge migration.
	Current knowledge migration methods transfer SD knowledge to the TD by aligning the distribution gap between domains.
	Such approaches deliver SD knowledge at a lower utilization rate.
	Actually, knowledge in FSCM exists in features, feature distributions, and overall category relationships.
	Foremost, high-quality transferable features are the most vital beginnings of knowledge transfer.
	Then, the distribution alignment migrates generic target knowledge in the global perspective.
	Building on this, the abundant category knowledge in the SD will refine the coarse generic migrated knowledge to yield fine-grained category-related transferable knowledge.
	Thus, it is plausible to believe that the multilevel process of knowledge transfer will yield more comprehensive and rich target representation for TD.
\end{itemize}

Here, we have developed a novel multi-level heterogeneous knowledge transfer network that establishes multilevel feature migration from FSCM data to measured data to better assist the task of SAR ATR.

\subsection{Contribution}
The proposed method, with FSCM and measured SAR images as inputs, achieves fuller knowledge migration from feature, distribution and category levels, respectively.
The main contributions of the proposed method are as follows:
\begin{enumerate}
	\item [1)] 
	FSCM data, which uses SC to model the target, describes the target essential information.
	Thus, this paper explores methods for knowledge migration from FSCM data to measured data.
	To the best of our knowledge, this is the first time that the FSCM are successfully employed to assist the measured SAR images target recognition via the deep learning framework.
	\item [2)] 
	There is significant heterogeneity between FSCM and measured SAR images. 
	For the purpose of knowledge migration from FSCM to measured data, we propose MHKT which migrates pure target information from the feature, sample and category levels, respectively.
	Crucially, the proposed method migrates FSCM knowledge completely to assist the SAR ATR task.
	\item [3)] The proposed method achieves state-of-the-art performance on two new datasets containing FSCM data and
	measured data. 
	Ablation experiments also demonstrated the necessity of multilevel feature migration.
\end{enumerate}

The remainder of the paper is organized as follows. In Section II, we review the related work. Section III describes the principles of FSCM data generation.
The overall architecture of the algorithm with the details of each module are described by Section IV.
Section V reports the experiment results and analysis.
Our conclusion and future work are presented in Section VI.

\section{Related Work}
\subsection{Forward Scattering Center Model}
In the high-frequency region, the backscattered echo of the target can be approximated as the sum of the responses of multiple SCs.
The SC model provides a concise and physically relevant description of the echo signal from a target, and is extensively applied in SAR ATR systems.
The solution of SC model parameters is divided into forward and inverse methods.
Inverse methods perform parameter estimation on measured image or echo data to obtain model parameters.
These methods derive model parameters that lack an explicit physical explanation and are hard to associate with target components.

To deal with the above issues, researchers from Electromagnetic Engineering Laboratory (EEL) at Wuhan University proposed a geometric model-based forward modeling method for complex target component-level SCs.
This approach establishes an excellent correspondence between the structure of the components and the model parameters.
Compared to the inverse method, the physical significance of the SCs from the forward method are explicit and are free for environment variations, such as background.

Given the mentioned advantages of FSCM, this type of simulation data has received widespread attention in the field of target recognition \cite{8315108}.
Some scholars \cite{8364125,7879812} have proposed a variety of matching criteria based on SC location and attribute parameters, which are founded on the advantages of local separability, and strong physical interpretability of the FSCM model.
Driven by the forward electromagnetic scattering model, the designed SAR ATR method is highly adaptable and can effectively improve the recognition performance.
However, these methods failed to address the more practical problem of limited training samples.

\subsection{Heterogeneous domain adaptation}
Unlike homogeneous data, the most obvious characteristics of heterogeneous data are that they have different data representations or dimensions.
That leads to the fact that the knowledge transfer framework of Fig. \ref{figI1} is no longer appropriate.

Lang et al. \cite{9784428} first explored a heterogeneous transfer learning approach in SAR ship recognition.
They utilized textual information from automatic identification systems (AIS) to extract geometric features of ships to aid classification tasks in the SAR image domain. 
AIS is in the form of text and includes both static (type, length, width, etc.) and dynamic (speed, position, etc.) information about the ship.
Their study demonstrated that the AIS information of ships can significantly improve the recognition performance with limited measured samples.
Moreover, they provide new solutions for SAR ATR under limited samples.

Additionally, heterogeneous domain adaptation (HDA) methods are the research focus in other fields.
HDA essentially solves the task of transferring knowledge between domains with various feature representations.
The state-of-the-art methods learn domain-invariant features by minimizing the divergence metric \cite{ZHU2019214, 10.1007/978-3-319-49409-8_35} or confusing the domain discriminator (i.e adversarial learning based methods) \cite{Li_2020,NEURIPS2018_ab88b157}.
Practically, the adversarial DA approach fulfils the concept of metric-based methods in a `black-box' manner.
Some representative metric methods include maximum mean discrepancy (MMD) \cite{ZHU2019214}, correlation alignment (CORAL) \cite{Sun2016DeepCC}, central moment discrepancy (CMD) \cite{zellinger2017central}, etc.
These methods align the marginal distributions by measuring the discrepancy between the various order moments of the distributions across the domains.
However, these methods migrate knowledge in a more general way and lack constraints on the discriminability of features.

On the other hand, adversarial learning based methods \cite{10.1145/3394171.3413995,pmlr-v70-arora17a} learn domain invariant features by fooling the domain discriminator, i.e., failing to determine whether instances come from SD or TD.
Recently, Long et al. \cite{NEURIPS2018_ab88b157} showed that a common adversarial domain adaptation framework does not guarantee that two domains are fully aligned even if the domain discriminator is completely obfuscated.
Furthermore, the instability associated with adversarial learning based methods are unavoidable regardless of the modifications.
 
Thus, we propose a new metric function that migrates more discriminative heterogeneous target knowledge.

\section{Solution of FSCM parameters}

\subsection{Attribute scattering center models}
Attribute scattering center (ASC) is a successful method for electromagnetic modelling of SAR targets.
FSCM is the forward method for the ASC.
The ASC model of the target can be expressed as:
\begin{equation}
	\begin{aligned}
		E^s(f, \varphi) = \sum_{i=1}^p & A_i\left(j \frac{f}{f_c}\right)^{\alpha_i} \sin c\left(k L_i \sin \left(\varphi-\varphi_i^{\prime}\right)\right)  
		\\ & \times \exp \left(-k c \gamma_i \sin \varphi\right) 
		\\ & \times \exp \left\{-j 2 k\left(x_i \cos \varphi+y_i \sin \varphi\right)\right\} 
	\end{aligned}
\end{equation}
where $\varphi$ and $f$ indicate azimuth and frequency. 
The sum of the backscatter field from the $p$ SCs forms the whole target backscatter.
$A_i$, $L_i$, ${\alpha}_i$, $\varphi_i^{\prime}$ and $(x_{i}, y_{i})$ represent the amplitude, the size of the local structure, frequency dependence factor, azimuth angle, and position parameters respectively.
These parameters characterise the SC features of the target.
Particularly, when $L_i=0$, the SC is defined as the local type; otherwise, the SC is distributed.
Given the three-dimensional properties of the real target, the FSCM estimates of the position parameters are also upgraded to three dimensions $(x_{i}, y_{i}, z_{i})$.

\subsection{Calculation of FSCM parameters}
The parametric modelling automation requires the completion of partitioning, ray tracing and parameter extrapolation, as shown in Fig. \ref{figI3}. 
The parameter calculation process is described below:
\begin{figure}[!t]
	\centering \includegraphics[width=3.0in]{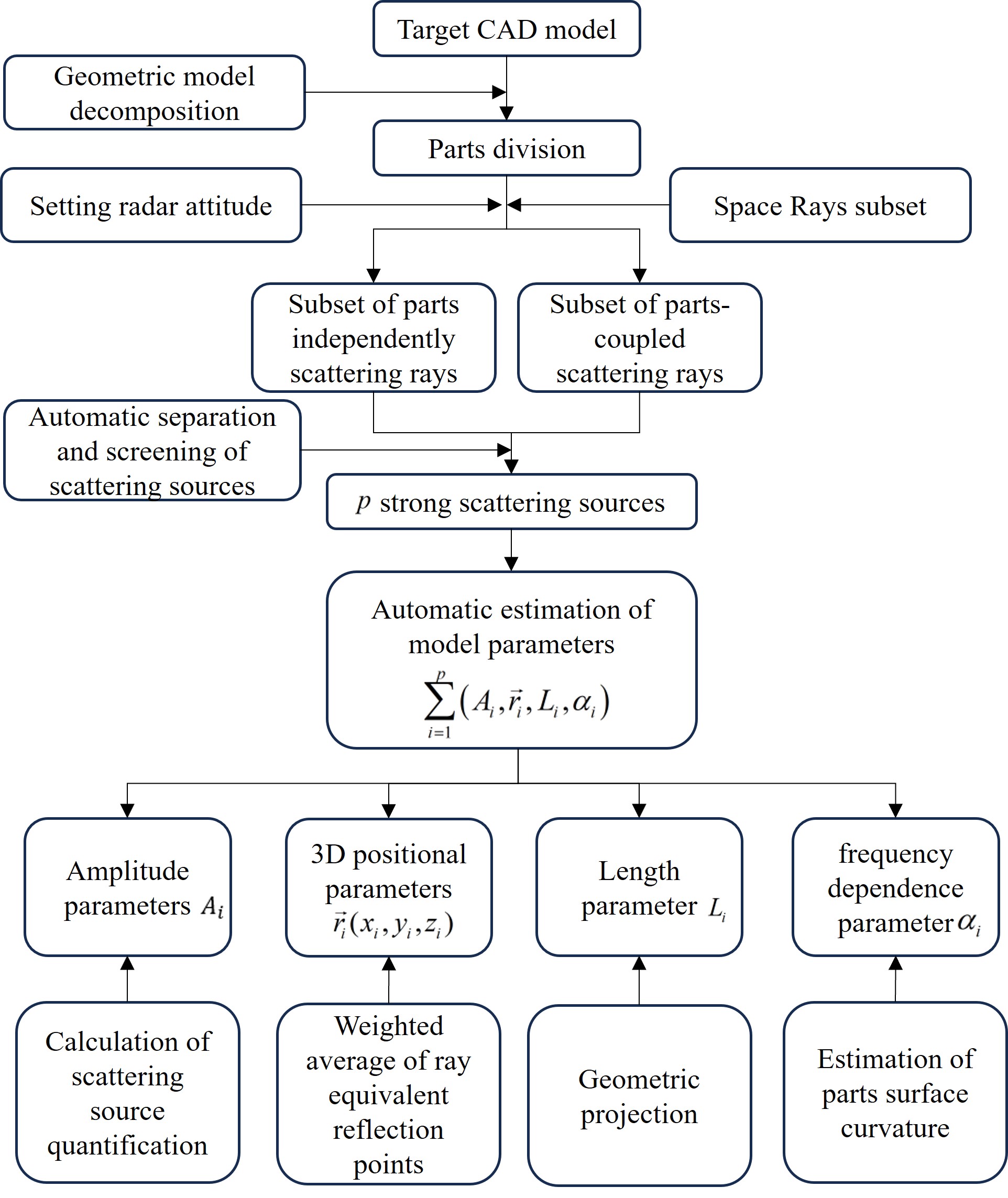}%
	\caption{Calculation process for FSCM parameters.}
	\label{figI3}
\end{figure}

1) Target scattering source separation and selection.
The geometric model is decomposed into a combination of multiple parts, firstly.
Then, the scattered field of the target is divide into the subset of intra-component and inter-component coupled scattered rays.
Each subset is a candidate scattering source.
The scattered contribution of each candidate is the superposition of all the ray fields in the subset.
Finally, the dominant strong scattering source is selected as the modelled SC.

2) Extrapolation of FSCM parameters: position, amplitude, length, and frequency dependence factor.
\begin{enumerate}
	\item [i)] Estimation of amplitude parameter $A$.
	Firstly, the magnitude of the scattered field for each ray is calculated.
	Summing all the ray fields from the SC will yield the its amplitude parameter.
	\item [ii)] Estimation of position parameter $(x, y, z)$.
	Ultimately, the positions are obtained by weighted averaging of the equivalent positions from all rays in the SC.
	Their weights are determined by the induced currents excited from the incident field on the target surface.
	\item [iii)] Estimation of length parameter $L$.
	Firstly, the structure type of the SC is evaluated.
	If each ray in the SC subset satisfies same-phase scattering, the SC can be evaluated as a distributed SC initially.
	Subsequently, the distance $L$ between the most distant equivalent reflection points is calculated.
	If $L \geq L_0$, the SC can be regarded as distributed, and the length parameter is $L$. 
	Else, it is identified as the local SC, and the length parameter is $0$.
	$L_0$ takes the value of
	\begin{equation}
		L_0=\frac{c}{2 f \sin \left(\frac{\Delta \varphi}{2}\right)}
	\end{equation}
	where $\Delta \varphi$ is the azimuth resolution.
	\item [iv)] Estimation of  frequency dependence parameter $\alpha$.
	It is closely related to the electromagnetic scattering mechanism and the curvature of the scattering structure.
	They are mainly classified as planar, hyperbolic or monoclinic surface, which corresponds to the frequency factor of 1, 0.5, and 0, respectively.
\end{enumerate}

\section{methodology}
\begin{figure*}[!t]
	\centering \includegraphics[width=6.5in]{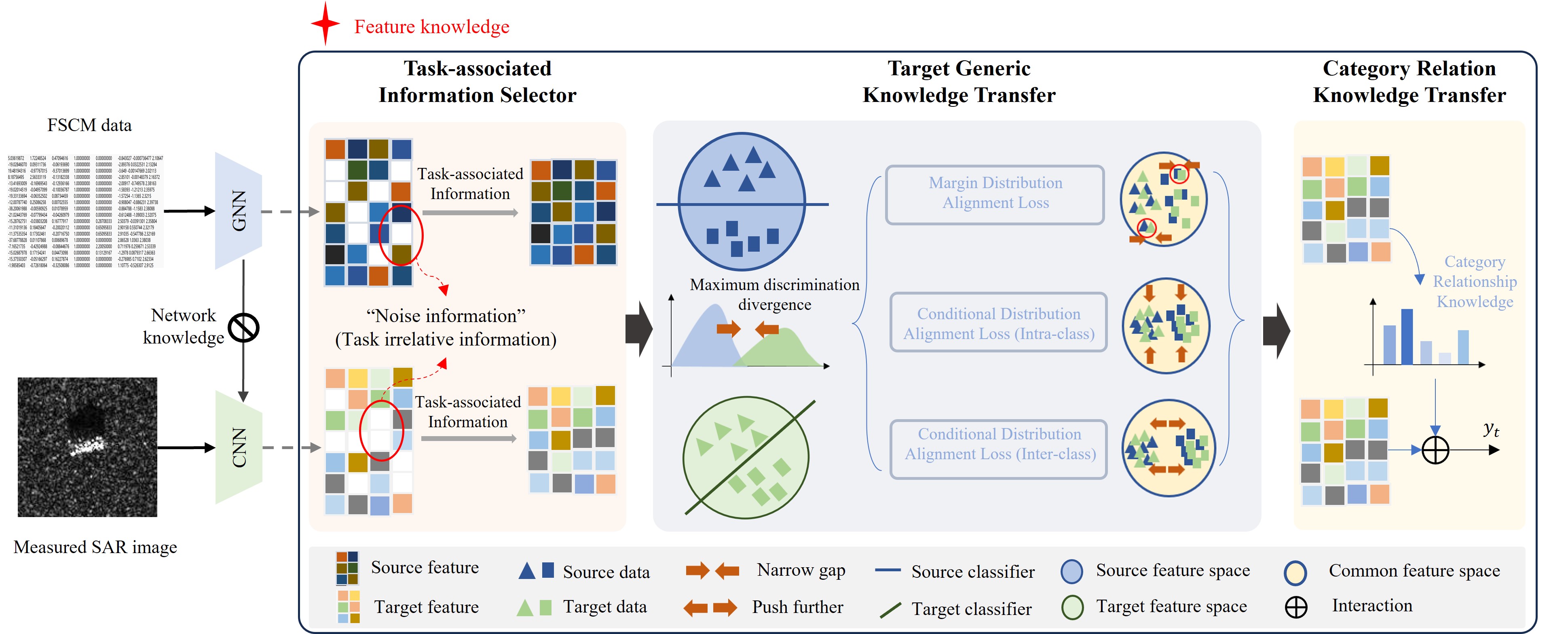}%
	\caption{The structure of multi-level heterogeneous knowledge transfer network on FSCM.}
	\label{figM1}
\end{figure*}

In this section, we mathematically model the vital problem to be addressed in this paper, firstly.
Then, the structure of the proposed modules and the corresponding roles are described in detail.
The overall structure of the proposed method is shown in Fig. \ref{figM1}.

\subsection{Problem modeling}
In this paper, we treat the measured SAR image as the TD and the FSCM data as the SD.
Here, a small number of labelled measured SAR images and a large number of FSCM data are utilized to train the network.
The SD and TD are represented as $D_s=\left\{\mathrm{X}_s, Y_s\right\}=\left\{\left(x_{s,i}, y_{s,i}\right)\right\}_{i=1}^{n_s}$ and $D_t=\left\{\mathrm{X}_t, Y_t\right\}=\left\{\left(x_{t,i}, y_{t,i}\right)\right\}_{i=1}^{n_t}$, separately.
$X$, $Y$ denote the data sets and labels, and $x$, $y$ are the sample and label of the instance.
$d_s$ and $d_t$, $n_s$ and $n_t$ mean the feature dimensions and sample size in SD and TD, respectively.
There exists $d_s \neq d_t$ and $n_t \textless n_s$.
Due to the domain shift, there exists gaps on the marginal and conditional distributions between the two domains, i.e., $P_s(x_s) \neq P_t(x_t)$, $Q_s(y_s|x_s) \neq Q_t(y_t|x_t)$.
Our purpose is to exploit SD sample knowledge in the TD classification task.

\subsection{Initial feature extraction}
\begin{figure}[!t]
	\centering \includegraphics[width=3.3in]{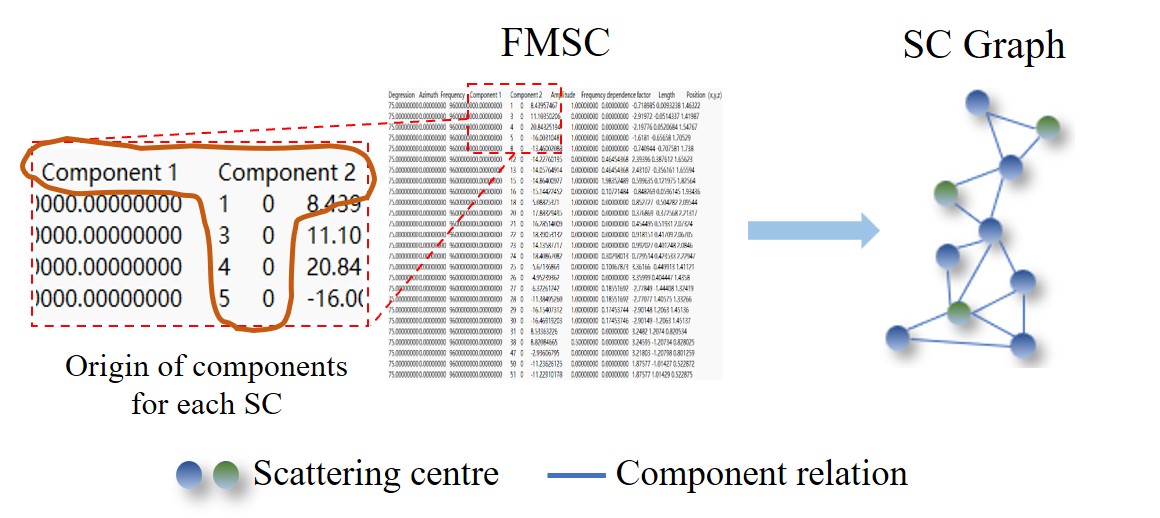}%
	\caption{The graph structure of the FSCM data is constructed based on the real target components.}
	\label{figG}
\end{figure}

Firstly, given that data from various domains have special forms of expression, we adopt different feature extraction methods to obtain the optimal performance.
FSCM data exhibit the discrete and disordered form. 
Thus, we draw on previous studies \cite{10137878} to model SCs with graph structure to extract structured electromagnetic scattering features (SESF).
Particularly, the greatest strength of the FSCM is the correspondence with the target physical components, thus the form of the graph structure can reflect the target geometry and the interactions among the components, as shown in Fig. \ref{figG}.
Subsequently, the graph neural network is employed to extract the 128-D SESF representation.
And for the measured SAR images, the AconvNet network is employed to extract 3136-D image domain features.
Evidently, although they describe the same target, there are significant differences in the feature dimensions, as listed in Table \ref{t2}.

\begin{table}[!t]
	\caption{Detailed information on heterogeneous features of FSCM data and SAR images}
	\renewcommand\arraystretch{1.7}
	\centering
	\label{t2}
	\setlength{\tabcolsep}{1.8mm}{
		\begin{tabular}{c c c c} \hline
			
			Dataset & Type of data & Feature &Dimension  \\ \hline
			FSCM   & Scattering center  & SESF  & 128\\
			SAR image  & image  & AconvNet & 3136   \\ 
			\hline 
		\end{tabular}
	}
\end{table}

\subsection{Task-associated information extraction}
\begin{figure}[!t]
	\centering
	\subfloat[]{\includegraphics[width=0.7in]{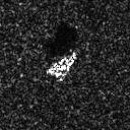}}%
	\label{Real SAR image}
	\hfil
	\subfloat[]{\includegraphics[width=0.6in]{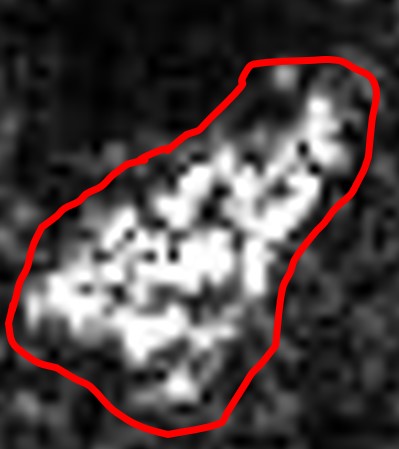}}%
	\label{ASC reconstructed SAR image}
	\hfil
	\subfloat[]{\includegraphics[width=0.7in]{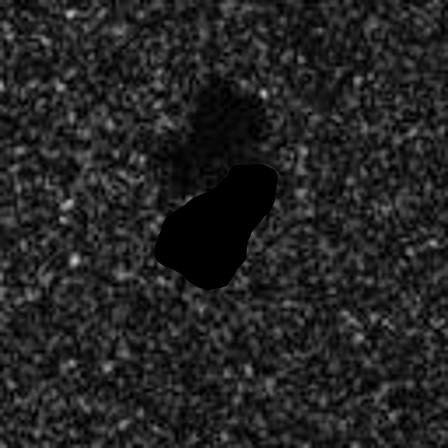}}%
	\label{ASC reconstructed SAR image}
	\caption{Comparison of different information on T72. (a) SAR image of the T72. (b) Regional demonstration of beneficial features in T72. (c) Regional demonstration of harmless features in T72.}
	\label{figI2}
\end{figure}

Obviously, the features from diverse domains contain information about the targets as well as task independence information, as shown in Fig. \ref{figI2}.
Aligning all feature thoughtlessly will lead to the negative migration phenomenon.
For this reason, we choose the advanced variational information bottleneck theory \cite{alemi2017deep} to decouple task-associated beneficial information.

\subsubsection{Variational information bottleneck theory}
Assume that the samples and labels are $X$ and $Y$, respectively.
The output feature $Z$ of the middle layer in the network is the result of encoding $X$, as determined by the encoder $P(Z|X;\Psi)$.
Thus, we can exploit Markov chains to reformulate the model of the classification task: $X \leftrightarrow Z \leftrightarrow Y$.
To facilitate the discussion, we treat $x$, $y$, $z$ as the specific instance.
Our goal is to make the new encoded features $Z$ maximize the information about the categories and compress the futile information in $X$ as much as possible.
The mutual information between $X$ and $Z$ as well as $Z$ and $Y$ can be exploited to model such process.
The objective function can be expressed as:
\begin{equation}
	max(I\left(Z, Y ; \Psi\right)-\beta I\left(X, Z ; \Psi\right))
\end{equation}
where $I(Z,Y;\Psi)$ indicates the mutual information between $Z$ and $Y$.
The mutual information $I(X, Z;\Psi)$ is from $Z$ and $X$.
And $\beta$ is the Lagrange multiplier.
In the first term, we leverage $Z$ to predict $Y$, which maintains the category information.
Second term encourages $Z$ to `forget' $X$ to compress the irrelevant information in $X$.

The modelling process is elegant and efficient, yet the computational process of mutual information is challenging.
We unfold the $I(Z,Y;\Psi)$ as follows:
\begin{equation}
	\begin{aligned}
		I(Z, Y; \Psi) & =\int d y d z p(y, z;\psi) \log \frac{p(y, z;\psi)}{p(y;\psi) p(z;\psi)}
		\\ & =\int d y d z p(y, z;\psi) \log \frac{p(y \mid z;\psi)}{p(y;\psi)}
	\end{aligned}
\end{equation}
Given that $p(y \mid z;\psi)$ is intractable, the $q(y \mid z; \phi)$ is employed as the variational approximation to $p(y \mid z;\psi)$.
$q(y \mid z; \phi)$ is a decoder which is update by parameter $\phi$.
Depending on the fact that the Kullback Leibler (KL) divergence is a nonnegative number, we obtain the following inequality:
\begin{equation}
	\begin{aligned}
		&KL[P(X \mid Z;\Psi), Q(Y \mid Z; \Phi)] \geq 0  \\ \Rightarrow 
		& \int d y p(y \mid z;\psi) \log p(y \mid z;\psi) 
		\\ & \geq \int d y p(y \mid z;\psi) \log q(y \mid z;\psi)
	\end{aligned}
\end{equation}
and hence
\begin{equation}
	\begin{aligned}
		I(Z, Y;\Psi) & \geq \int d y d z p(y, z;\psi) \log \frac{q(y \mid z;\phi)}{p(y;\psi)} \\
		& =\int d y d z p(y, z;\psi) \log q(y \mid z;\phi) 
		\\& \quad -\int d y p(y;\psi) \log p(y;\psi) \\
		& =\int d y d z p(y, z;\psi) \log q(y \mid z;\phi)+H(Y)
	\end{aligned}
\end{equation}
where the entropy of the label $H(Y)$ is irrelevant to our optimization process which can be ignored.
Additionally, since the $p(y, z;\psi)=\int d x p(x, y, z;\psi)=\int d x p(x;\psi) p(y \mid x;\psi) p(z \mid x;\psi)$, we rewrite the (6) as
\begin{equation}
	\begin{aligned}
		I\left(Z, Y ; \Psi, \Phi\right) \geq  \quad \int 
		d{x} d{z} d{y} & p\left(x;\psi \right) p\left(y \mid x;\psi \right) \\
		& \times p\left(z \mid x;\psi \right) \log q\left(y \mid z;\phi \right)
	\end{aligned}
\end{equation}

Subsequently, we unfold the second term $I (X,Z;\Psi)$ in (3) as follows
\begin{equation}
	\begin{aligned}
		I\left(X, Z ; \Psi\right)= & \int d{x} d{z} p\left(z, x;\psi \right) \operatorname{logp}\left(z \mid x;\psi \right) \\
		& -\int d{z} p\left(z;\psi \right) \operatorname{logp}\left(z;\psi\right)
	\end{aligned}
\end{equation}
It is also extremely hard to compute the marginal distribution of $Z$.
We let $r(z)$ be the variational approximation of $p(z)$.
Since KL divergence greater than or equal to $0$, an upper bound for $I(X, Z)$ can be obtained, as follows:
\begin{equation}
	I(X, Z) \leq \int dx dz p(x) p(z \mid x) \log \frac{p(z \mid x)}{r(z)}
\end{equation}

Combining inequalities (7) and (9), we rewrite the objective function (3) as
\begin{equation}
	\begin{aligned}
		I(&Z, Y; \Psi)-\beta I(X, Z; \Psi) \geq L_{LB} \\
		= &\int d x d y d z p(x;\psi) p(y \mid x;\psi) p(z \mid x;\psi) \log q(y \mid z_;\phi) \\
		& -\beta \int d x d z p(x;\psi) p(z \mid x;\psi) \log \frac{p(z \mid x;\psi)}{r(z)} \\
	\end{aligned}
\end{equation}
Thus, we obtain the lower bound $L_{LB}$ of (3).
Practically, we can approximate $p(x,y)=p(x)p(y \mid x)$ by the empirical data distribution $p(x,y)=\frac{1}{N} \sum_{i=1}^{N} \delta_{x_{i}}(x) \delta_{y_{i}}(y)$.
$N$ is the number of samples.
Hence, the lower bound $L_{LB}$ can be rewritten as
\begin{equation}
	\begin{aligned}
		L_{LB} \approx \frac{1}{N} \sum_{i=1}^{N} [\int 
		& dz p(z \mid x_n; \psi ) \log q(y_n \mid z;\phi) \\
		& \quad -\beta p(z \mid x_n;\psi) \log \frac{p(z \mid x_n;\psi )}{r(z)}] \\
	\end{aligned}
\end{equation}

According to the variational encoder principle, we define the encoder as a Gaussian distribution $p(z \mid x)=N\left(z \mid f^{\mu}(x), f^{\sigma}(x)\right)$. 
$f$ denotes the information bottleneck layer.
$\mu$ and $\sigma$ denote mean and variance, respectively.
The reparameterization trick is exploited to write $p(z \mid x) d z=p(\epsilon) d \epsilon$, where $z=f(x, \epsilon)$ is a deterministic function of $x$ and the Gaussian random variable $\epsilon$.
Thus, we can obtain the following loss function and try to minimize it:
\begin{equation}
	\begin{aligned}
		{L}_{\text {TAIS}}=& \frac{1}{N} \sum_{i=1}^{N}   \mathbb{E}_{\epsilon \sim p(\epsilon;\psi)}\left[-\log q\left(y_{i} \mid \mathbb{F}\left(x_{i}, \epsilon;\phi\right)\right)\right] \\
		& \quad  \quad +\beta K L\left[p \left(z_{i} \mid x_{i};\psi\right), r\left(z_{i}\right)\right]
	\end{aligned}
\end{equation}
The objective of the TAIS module is to minimize this loss function.

\subsubsection{Task-associated information selector}
\begin{figure}[!t]
	\centering \includegraphics[width=2.5in]{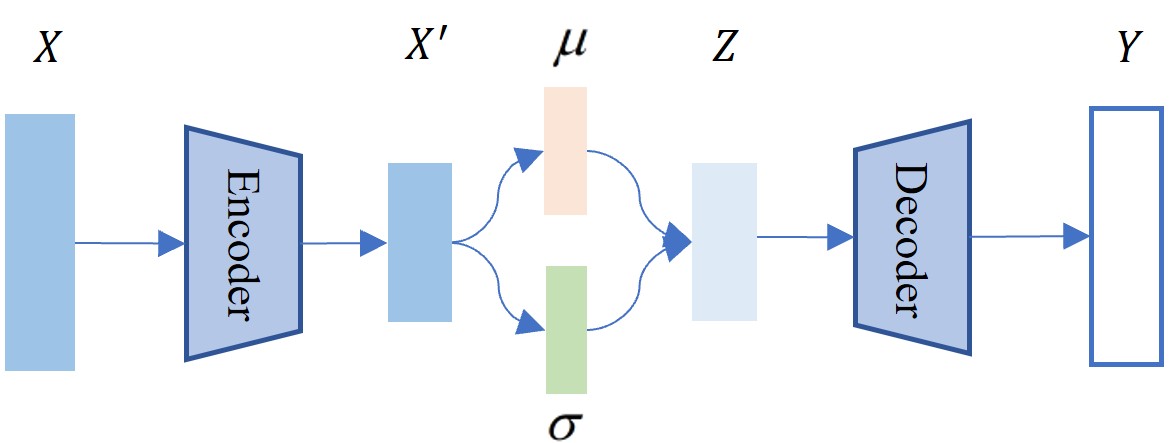}%
	\caption{Structure of task-related information selector.}
	\label{figM2}
\end{figure}

The specific structure of the TAIS module is shown in Fig. \ref{figM2}.
After encoding by the encoder $P(X, Z; \Psi)$, the new feature ${X}^{'}$ is generated.
Subsequently, we compute the mean and variance of the beneficial information in ${X}^{'}$ and reconstruct the beneficial information $Z$.

In summary, the TAIS module provide the following strengths:
\begin{itemize}
	\item [i)] Enhance the correlation between feature and category information and compress irrelevant information.
	\item [ii)] Implicitly align the feature distributions between the two domains.
	Task-related information in SD and TD is augmented, simultaneously. 
	The formula is shown below, 
	\begin{equation}
		X_s \leftrightarrow Z_s \leftrightarrow Y \leftrightarrow  Z_t \leftrightarrow X_t. 
	\end{equation}
	Evidently, the two domains are moving closer to the category information.
\end{itemize}

\subsection{Targeted generic knowledge transfer}

In this subsection, we present the new objective function to measure the inter-domain distributional gaps.
This metric function aligns the marginal and conditional distributions while increasing the intra-class density and pushing away the inter-class distance.

\subsubsection{Feature projection}
Subsequently, it is desirable to project these new features into the common feature space firstly, and then measure the inter-domain gaps, as follows.
	\begin{align}
		{u_s}=W_s z_s \nonumber \\
		{u_t}=W_t z_t 
	\end{align}

\subsubsection{Maximum discrimination divergence}
The purpose of DA in this paper is to minimize the MDD, which can be expressed as follows:
\begin{equation}
	\begin{aligned}
		\operatorname{MDD}(U_s, U_t)=& \mathbb{E}_{U_s \sim P_s, U_t \sim P_t}\left[\left\|U_s-U_t\right\|_2^2\right] \\
		& + \mathbb{E}_{U \sim C, U^{\prime} \sim C}\left[\left\|U- \widetilde{U}\right\|_2^2\right] +\mathbb{E}_{U \sim C, \widetilde{U} \sim C}\left[ U\widetilde{U}\right]
	\end{aligned}
\end{equation}
where $U=[U_s,U_t], U \sim C$, denotes the common feature space that includes samples from all domains.
$\widetilde{U}$ denotes a copy of $U$.
Evidently, the $\operatorname{MDD}$ consists of three important components.
The first measures the gap of the inter-domain marginal distribution.
The second term aligns the conditional distribution by incorporating categorical information.
Moreover, this term preserves the discriminative structure information of the labelled samples by constraining the intra-class distance.
Additionally, samples from different categories should have minimal similarity in the common feature space.
We adopt the dot product between two samples in a common space to measure the dissimilarity between them.
The smaller the result, the lower the similarity, which ensures the learnability of the objective function, as shown in the third part of (15).

Subsequently, we hypothesise that there are $n_s$, $ n_t$ samples in SD and TD respectively, which can be denoted as $\left\{\left(u_{s, i}, y_{s, i}\right)\right\}_{i=1}^{n_s}$ and $\left\{\left(u_{t, j}, y_{t, j}\right)\right\}_{i=1}^{n_t}$.
And $\left\{\left(u_i, y_i\right)\right\}_{i=1}^{n_s+n_t}$ express the summary of the SD and TD samples.
Then (15) can be reformulated as
\begin{equation}
	\begin{aligned}
		\operatorname{MDD}(U_s, U_t)  = & \frac{1}{n_s n_t} \sum_{i, j}^{n_s, n_t}\left\|u_{s, i}-u_{t, j}\right\|_2^2  + \frac{1}{m_1} \sum_{y_{i}=y_{j}}\left\|u_{i}-u_{j} \right\|_2^2  \\
		& + \frac{1}{m_2} \sum_{y_{i} \neq y_{j}} u_{i}u_{j}
	\end{aligned}
\end{equation}
where $i$, $j$ indicate the sample index.
$m_1$ and $m_2$ represent the number of samples of the same and distinct categories in the common feature space, respectively.
Given that the second term measures the variability between samples with the same category, we rewrite that as follows:
\begin{equation}
	\frac{1}{m_1} \sum_{i, j=1}^{n_s+n_t}\left\| {u}_i- {u}_j\right\|_F^2 {W}_{i j}
\end{equation}
where ${W}_{i j}$ indicates the similarity matrix in the common feature space, which can be defined as
\begin{equation}
	{W}_{i j}= \begin{cases}1, & \text { if } y_i=y_j \\ 0, & \text { otherwise. }\end{cases}
\end{equation}
Then, we unfold (17) to obtain a new representation as
\begin{equation}
	\frac{2}{m_1} \left( \sum_{j=1}^n {u}_j^T {u}_j \widetilde{{D}}_{j j}- \sum_{i, j=1}^n {u}_i^T {u}_j {W}_{i j} \right)=\frac{2}{m_1} \operatorname{tr}\left(\mathbf{U} \mathbf{L}_H \mathbf{U}^{\mathrm{T}}\right)
\end{equation}
where $\mathbf{L}_H=\widetilde{\mathbf{D}}- \mathbf{W}$ is the Laplacian matrix. $\widetilde{{D}}_{j j}=\sum_{i} {W}_{i j}$ and the other elements in $\widetilde{\mathbf{D}}$ are 0.

By minimizing Eq. (19), samples with the same label (whether from the SD or TD) are close to each other in the common feature space.
That enhances the similarity of cross-domain samples with the same labels, narrows the conditional distributions of the two domains, and realizes the transfer of effective knowledge.

The third term analyses the correlation between the samples to measure the dissimilarity, which can be rewritten as
\begin{equation}
	\left\|\mathbf{B} \odot\left(\mathbf{u}^T \mathbf{u}\right)\right\|_F^2=2 \sum_{y_i \neq y_j}\left({u}_i^{\mathrm{T}} {u}_j\right)^2 .
\end{equation}
where $\mathbf{B}$ is the dissimilarity matrix, expressed as follows:
\begin{equation}
	{B}_{i j}= \begin{cases}1, & \text { if } y_i \neq y_j \\ 0, & \text { otherwise. }\end{cases}
\end{equation}
From the definition of $\mathbf{B}$ and $\mathbf{W}$,we can find that $\mathbf{B}+\mathbf{W}=\mathbf{I}$, i.e., $\mathbf{B}$ and $\mathbf{W}$ can form an all-ones matrix $\mathbf{I}$.
Similarly, samples (regardless of SD or TD) with different labels in the common space can be pushed away from each other by minimizing (20).
This term reinforces the separability among classes and transfers the discriminative knowledge from the SD to the TD efficiently.

\begin{figure}[!t]
	\centering
	\subfloat[]{\includegraphics[width=1.2in]{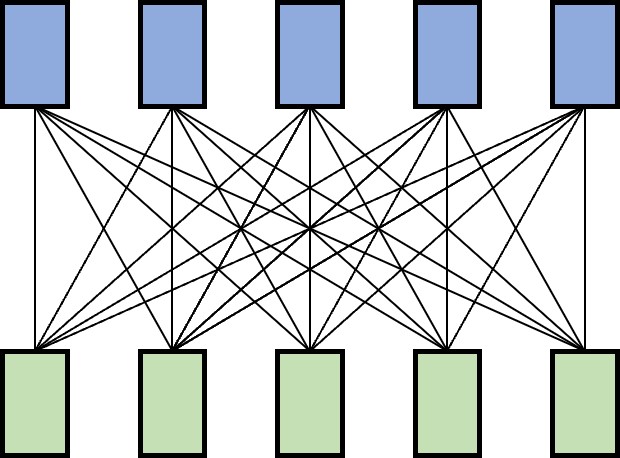}}%
	\label{Real SAR image}
	\hfil
	\subfloat[]{\includegraphics[width=1.2in]{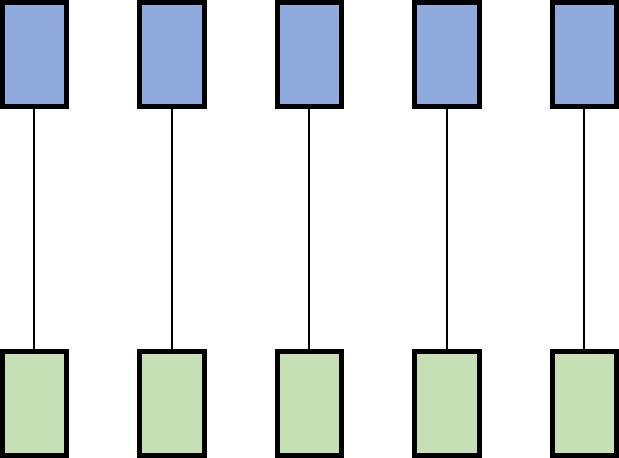}}%
	\label{ASC reconstructed SAR image}
	\caption{Computation of inter-domain marginal distributions. (a) Iteration over all samples. (b) Simplified version in this paper.}
	\label{figM3}
\end{figure}

\begin{figure}[!t]
	\centering \includegraphics[width=3.3in]{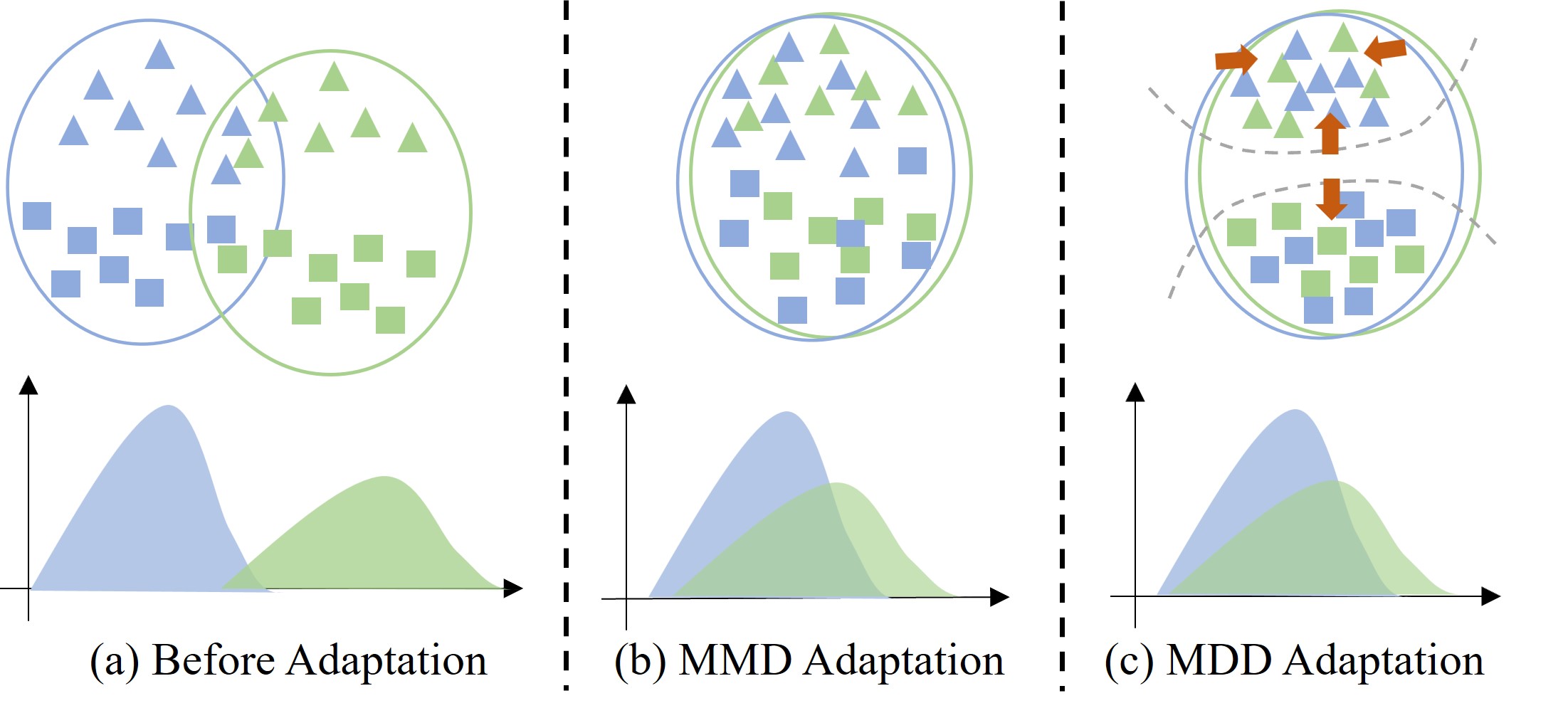}%
	\caption{Adaptation results for various metric functions.}
	\label{figM4}
\end{figure}

Additionally, the first term calculates the distance for pairwise samples.
Deep networks are generally trained in batches, where we receive a fraction of the entire dataset in each batch.
Thus, to facilitate the calculation of MDD, we only calculate the distance between the corresponding locations of the SD and TD in the batch, covering all cross-domain samples through successive iterations, instead of visiting all samples at once.
Fig. \ref{figM3} illustrates the computational principle of the modified first term.
Then, we rewrite (16) as
	\begin{align}
		L_{MDD} &=\frac{1}{n_s n_t} \sum_{i, j}^{n_s, n_t}\left\|u_{s, i}-u_{t, j}\right\|_2^2+\frac{1}{m_1} \sum_{y_{i}=y_{j}}\left\|u_{i}-u_{j} \right\|_2^2 \nonumber \\
		& \quad +\frac{1}{m_2} \sum_{y_{i} \neq y_{j}} u_{i}u_{j}  \nonumber \\
		& = \frac{1}{n_b} \sum_{i, j}^{n_b}\left\|u_{s, i}-u_{t, j}\right\|_2^2 + \frac{2}{m_1}\operatorname{tr}\left(\mathbf{U} \mathbf{L}_H \mathbf{U}^{\mathrm{T}}\right) \nonumber  \\
		& \quad + \frac{1}{2m_2} \left\|\mathbf{B} \odot\left(\mathbf{U}^T \mathbf{U}\right)\right\|_F^2
	\end{align} 
	where $n_b$ denotes the number of samples in the batch.
	Ultimately, the result of the MDD is shown in Fig. \ref{figM4}.


\subsection{Category relation knowledge transfer}
Despite the huge heterogeneity in the SD and TD, classifiers should generate similar category probability distribution for the same category from any domains.
This probability distribution reflects the correlation information among categories, which is a kind of implicit semantic knowledge.
Given soft labels contain a lot of valuable category relevance information, we exploit the SD data to construct soft labels about the semantics of categories, as follows below.
\begin{equation}
	q^{(k)}=\frac{1}{{n_s}^{(k)}} \sum_{\boldsymbol i \in n_s^{(k)}} \operatorname{softmax}\left(f_c\left(\boldsymbol{u}_{s,i}\right)\right)
\end{equation}
where ${n_s}^{(k)}$ denotes the number of samples with category $k$ in SD.
$f_c$ means the shared classifier.
A labelled target instance can fine-tune the target network with soft-label to transfer semantic associations from the SD to the TD.
Hence, under the supervision of the learnt soft-label, the corresponding loss can be calculated as
\begin{equation}
	{L}_{soft}\left(U_L, Y_L\right)=-\frac{1}{n_l} \sum_{\boldsymbol{u}_i \in U_L, y_i \in Y_L} q^{\left(y_i\right)^{\top}} \log \boldsymbol{p}_i
\end{equation} 
where $p_i$ is the probabilistic output for labeled target instance $x_i$ , and $p_i = softmax (f_c(x_i))$.
$ \left\{ U_L,Y_L \right\}$ represent the labelled target instances.
To conclude, we further consider the supervised loss of labeled target data and define the implicit semantic correlation loss as
\begin{equation}
	{L}_{CRKT}=(1-\alpha) {L}_{CE}\left(U_L, {Y}_L\right)+\alpha {L}_{soft}\left(U_L, {Y}_L\right)
\end{equation}
where $L_{CE}$ and $\alpha$ denote the cross-entropy loss function and balancing hyperparameter, individually.
And, $\alpha$ takes values in the range of $\left( 0, 1 \right)$.

\subsection{Overall Formulation and Optimization}
In summary, the target knowledge in FSCM is migrated exhaustively via a multilevel migration strategy, which from features, samples, and categories levels.
Specifically, more pure and discriminative target information is migrated through the constraints of the TAIS, TGKT and CRKT modules.
To this end, we represent the overall objective function as the following format:
\begin{equation}
	\begin{aligned}  
	L&={\lambda}_1 L_{TAIS}+ {\lambda}_2 L_{MDD}+L_{CRKT}
\end{aligned}
\end{equation}
where ${\lambda}_1$ and ${\lambda}_2$ are employed to balance the effects of TAIS and MDD on the model.

\section{Experimental results and analysis}

\subsection{Dataset}
\subsubsection{Sim2Mstar dataset}
Sim2Mstar dataset consisting of FSCM and Moving and Stationary Target Acquisition and Recognition (MSTAR) \cite{10.1117/12.242059} is constructed, in this paper.

\begin{figure}[!t]
	\centering \includegraphics[width=2.2in]{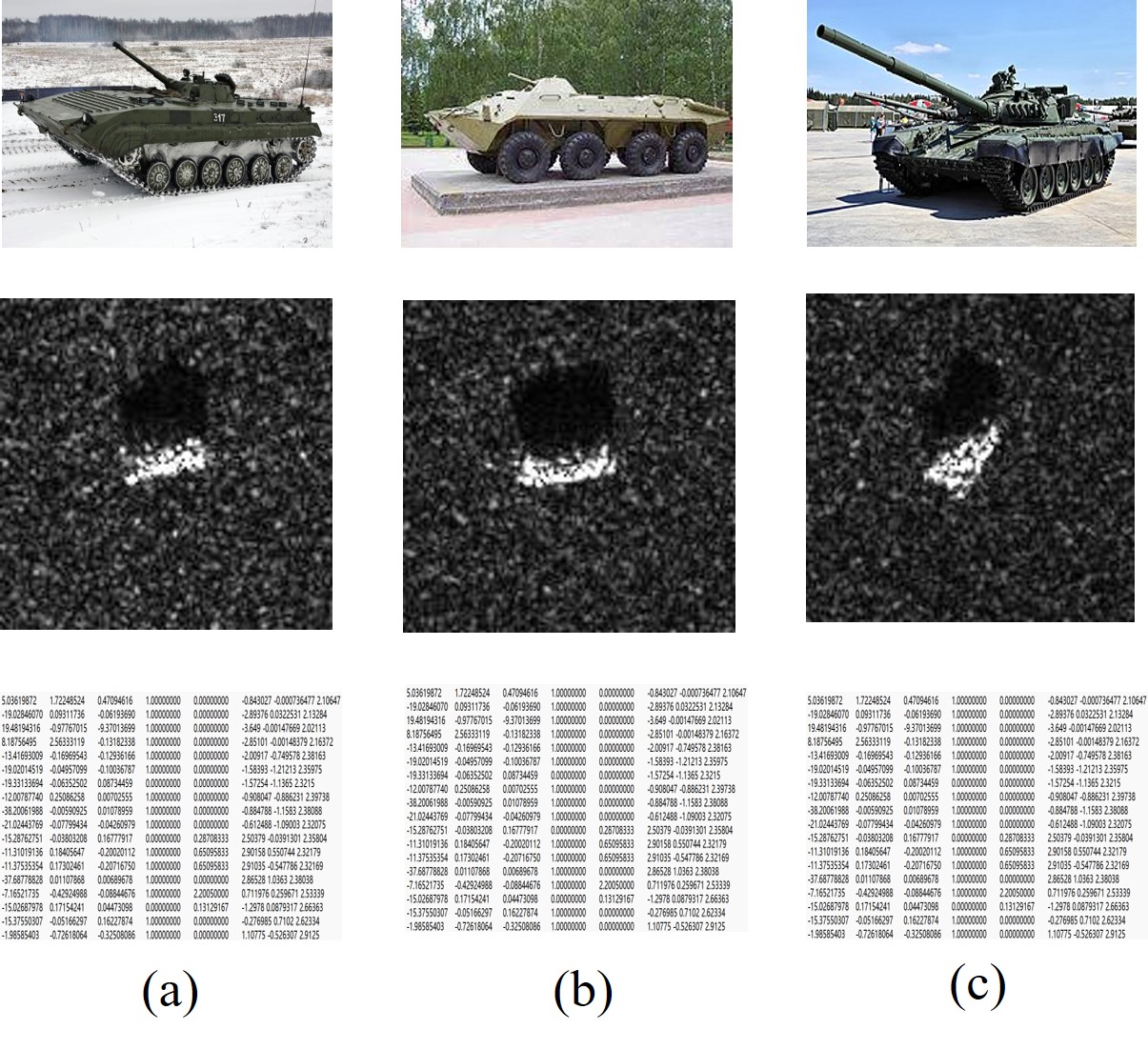}%
	\caption{Example of a new data set. The first row represents the optical image display of the different targets. The last two rows are the MSTAR measured SAR image and the simulated SCs corresponding to this target, respectively. (a) BMP2. (b) BTR70. (c) T72.}
	\label{fig51}
\end{figure}

The MSTAR dataset, sponsored by the US Department of Defense Advanced Research Projects Agency and the Air Force Research Laboratory, serves to evaluate the performance of the methods with completeness, diversity and standardisation.
The dataset contains SAR images with different configurations (e.g., type, azimuth, and radar depression angle).
In this paper, the classical targets: BMP2, BTR70, T72 are selected for recognition.
Fig. \ref{fig51} illustrates the optical images, SAR images and corresponding FSCM data for the three targets.
In the training phase, the SD and TD data are: the FSCM data and MSTAR images at $17^{\circ}$ depression angle (i.e., measured-train), respectively.
The test set data is MSTAR images at $15^{\circ}$ depression angle (i.e., measured-test), as shown in Table \ref{t1}.

\begin{figure}[!t]
	\centering \includegraphics[width=2.3in]{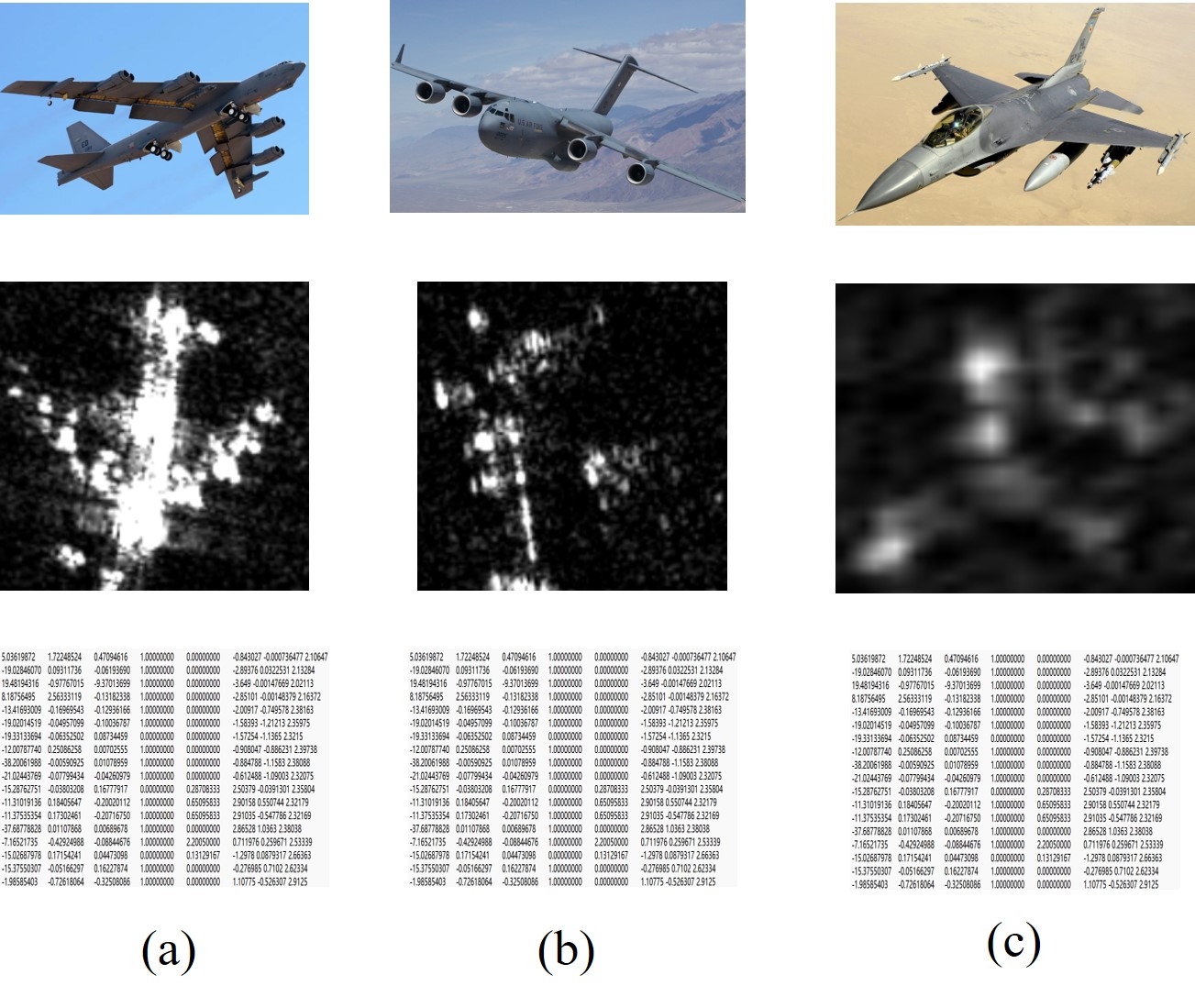}%
	\caption{Example of a new data set. The first row represents the optical image display of the different targets. The last two rows are the MSTAR measured SAR image and the simulated SCs corresponding to this target, respectively. (a) A. (b) B. (c) C.}
	\label{fig5a}
\end{figure}

\begin{table}[!t]
	\caption{Detailed information about the data in the Sim2Mstar dataset}
	\renewcommand\arraystretch{1.4}
	\centering
	\label{t1}

	\setlength{\tabcolsep}{2.6mm}{
		\begin{tabular}{c c c c c} \hline
			\multirow{2}{*}{Dataset} &\multirow{2}{*}{Depression}  &\multicolumn{3}{c}{Number} \\
			\cline{3-5}
			& & BMP2 & BTR70 & T72  \\ \hline
			FSCM   & $17^{\circ}$ & 360& 360 & 360 \\
			Measured-train  & $17^{\circ}$ & 233  & 233  & 232   \\ 
			Measured-test  & $15^{\circ}$  & 195  & 196  & 196   \\ 
			\hline 
		\end{tabular}
	}
\end{table}

\begin{table}[!t]
	\caption{Detailed information about the data in the Sim2Air dataset}
	\renewcommand\arraystretch{1.4}
	\centering
	\label{tF}

	\setlength{\tabcolsep}{2.6mm}{
		\begin{tabular}{c c c c c} \hline
			\multirow{2}{*}{Dataset} &\multirow{2}{*}{Depression}  &\multicolumn{3}{c}{Number} \\
			\cline{3-5}
			& & A & B & C  \\ \hline
			FSCM   & $17^{\circ}$ & 360& 360 & 360 \\
			Measured-train  & $17^{\circ}$ & 60  & 60  & 60   \\ 
			Measured-test  & $15^{\circ}$  & 73  & 73  & 73   \\ 
			\hline 
		\end{tabular}
	}
\end{table}

\subsubsection{Sim2Air dataset}

Sim2Air dataset includes the SAR aircraft image dataset and the corresponding FSCM data.
Specifically, the dataset covers three categories of aircraft.
Fig. \ref{fig5a} shows the optical images, SAR images and corresponding FSCM data for the three targets.
Similarly, the SD is the FSCM data of the aircraft, while the TD contains a small amount of measured SAR imagery, as shown in Table \ref{tF}.

It is worth mentioning that, given the richness and completeness of the Sim2Mstar dataset, we perform parametric analyses and ablation experiments on it.
Moreover, we validate the generalization of the proposed method on Sim2Air with slightly poorer data quality.

\subsection{Experiments on hyperparameters}
The multiple losses involved in the proposed method constrain the network learning process as depicted in (26).
Evidently, the impact of various losses are balanced by $\lambda_1$ and $\lambda_2$.
Additionally, the participation extent of the SD category relationship information contained in $L_{CRKT}$ is controlled by the crucial parameter $\alpha$ in (25).
It is essential to explore the effect of their diverse values on the network performance.
Specifically, we randomly select 10 measured samples for each class in Sim2Mstar for training.

\begin{figure}[!t]
	\centering \includegraphics[width=2.8in]{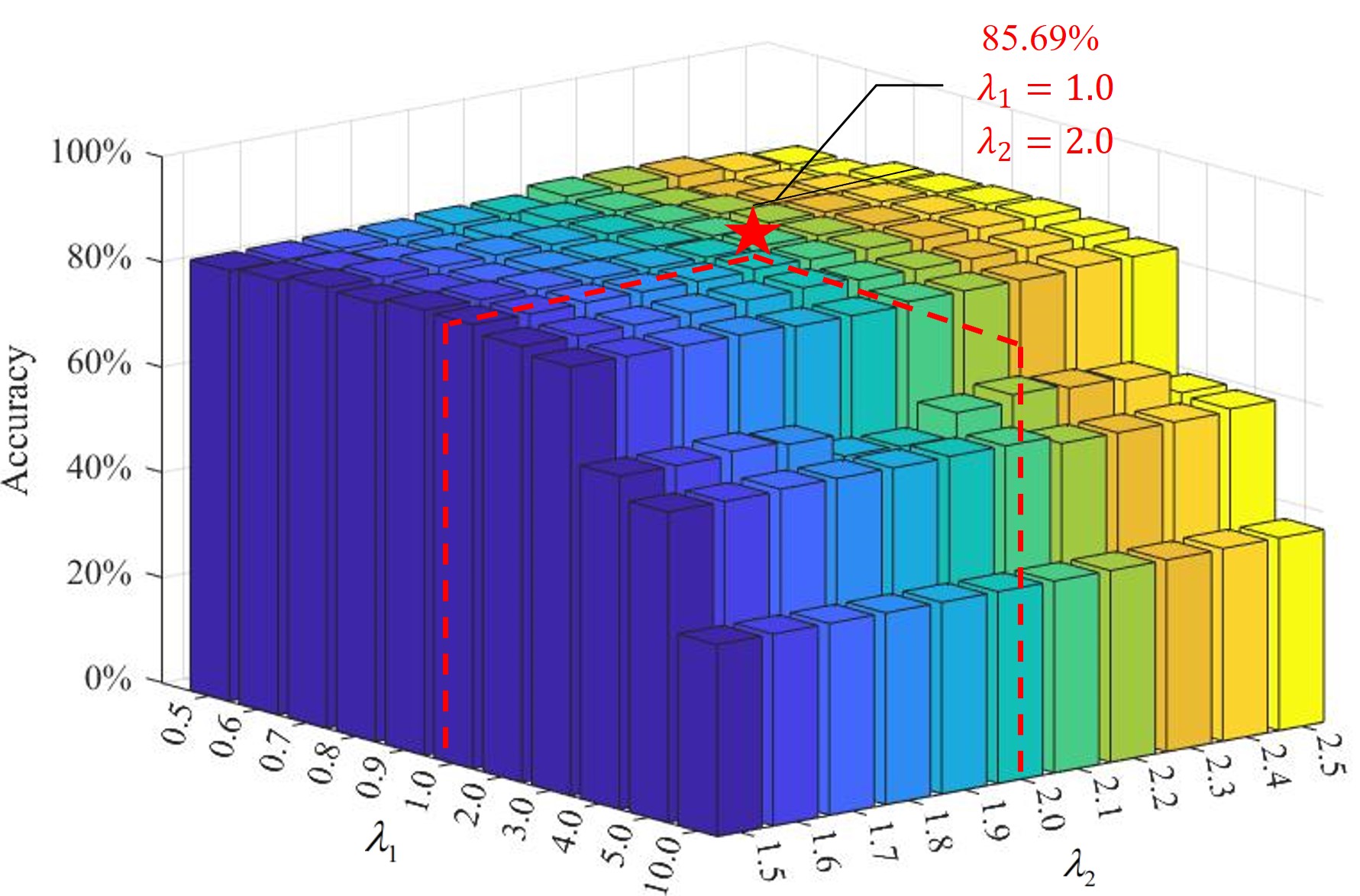}%
	\caption{Impact of $\lambda_1$ and $\lambda_2$ on network performance. 10 randomly selected samples from each class in the measured-train.}
	\label{figE2}
\end{figure}

\begin{figure}[!t]
	\centering \includegraphics[width=3.0in]{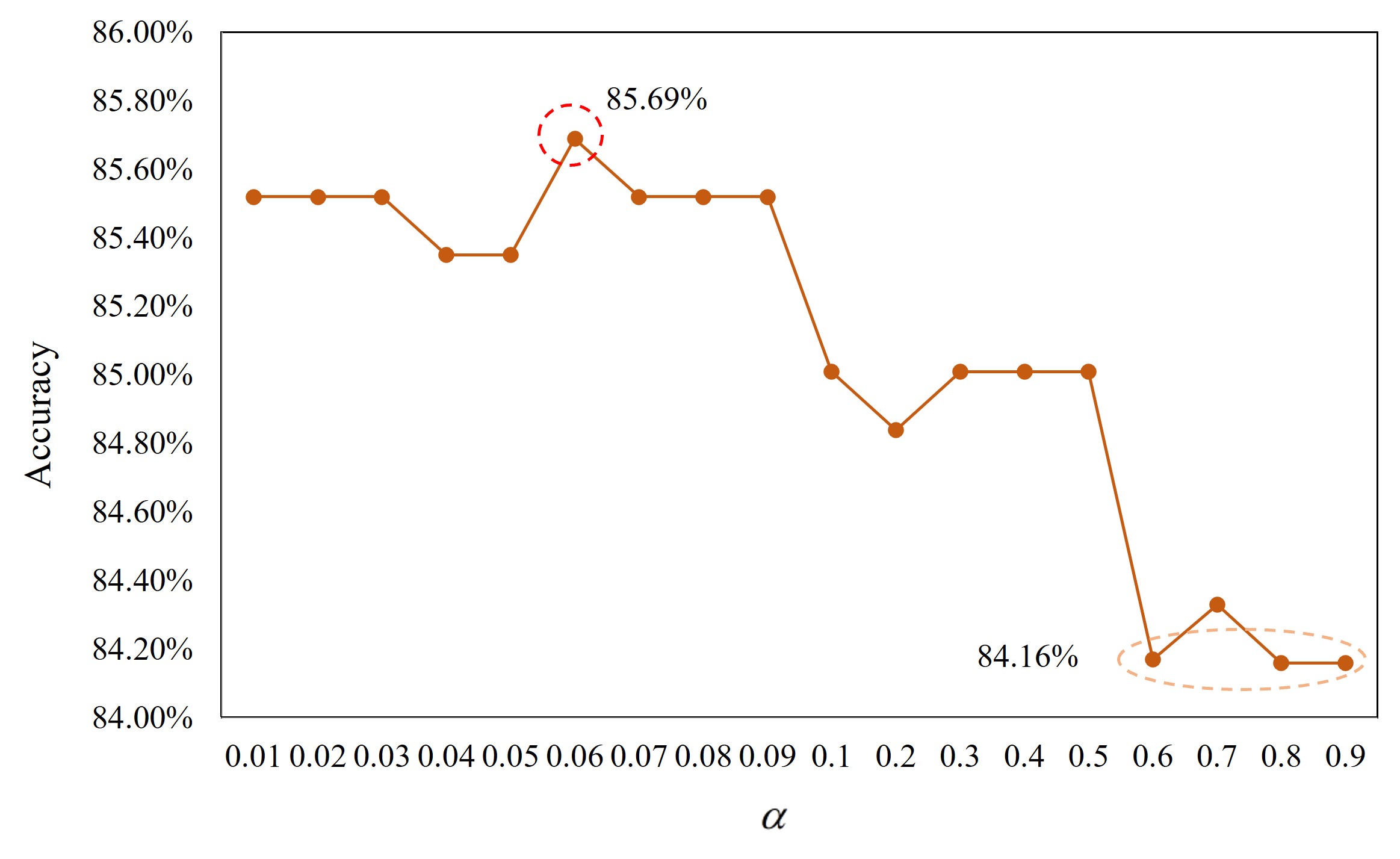}%
	\caption{Impact of $\alpha$ on network performance. 10 randomly selected samples from each class in the measured-train.}
	\label{figEa}
\end{figure}

\subsubsection{Effect of $\lambda_1$ and $\lambda_2$ on Performance}
Fig. \ref{figE2} presents the experiment results.
Apparently, the proposed method is robust to parameter variations.
A peak recognition accuracy of 85.69\% is indicated when $\lambda_1$ and $\lambda_2$ take the values of 1.0 and 2.0, respectively.
Especially, it is insensitive to changes in $\lambda_2$.
And, when the value of $\lambda_1$ is set within 3.0, the recognition accuracy changes insignificantly and stays above 80\%.
As the value of $\lambda_1$ increases further, the recognition accuracy drops off in a precipitous manner.
Since $\lambda_1$ directly controls TAIS module, when $\lambda_1$ grows to a certain level, the network moves from the original filtering of futile information to the further compression of task-associated information.
The reduction in beneficial knowledge inevitably leads to a decrease in recognition performance.
Notably, the recognition performance remains stable against changes in $\lambda_2$, in this situation.

\subsubsection{Effect of $\alpha$ on Performance}
$\alpha$ balances the contribution of simulation data labelling knowledge.
Fig. \ref{figEa} illustrates the corresponding recognition results when $\alpha$ takes different values.
Obviously, the recognition performance reaches the peak when $\alpha$ takes the value of 0.06.
As $\alpha$ increases further, the performance program weakly tends to decrease.
When the values are 0.6, 0.8 and 0.9, the performance drops by 1.53\% and the recognition accuracy reaches 84.16\%.
Thus, the performance of the proposed method is relatively stable with $\alpha$ changes.
This phenomenon indicates that the CRKT module steadily migrates category semantic knowledge from FSCM to the measured data.

\subsection{Ablation experiment}

In this section, we maintain consistency with the data setting in the Experiments on hyperparameters.

\begin{table}[!t]
	\renewcommand\arraystretch{1.4}
	\caption{Ablation experiments on different modules of the network. 10 randomly selected samples from each class in the measured-train.\label{t3}}
	\centering
	\scalebox{1}{
		\setlength{\tabcolsep}{1.6mm}{\begin{tabular}{c c c c c }
				\hline
				Modules     & TAIS     & TGKT    & CRKT      & Overall Accuracy \\ 
				\hline
				Target Only       & -       & -      & -    & 77.78\% \\
				\hline
				\multirow{4}{*}{Individual Modules (Ours)}   
				& $\times$   & $\times$      & $\times$    & 75.64\% \\
				& $\checkmark$   & $\times$      & $\times$    & 82.62\% \\
				&$\times$   & $\checkmark$  & $\times$     & 81.37\%\\
				&$\times$   & $\times$      & $\checkmark$     &  80.41\% \\
				\hline
				\multirow{3}{*}{Module coupling (Ours)}   
				&$\checkmark$   & $\checkmark$  & $\times$    & 83.13\%\\
				&$\times$   & $\checkmark$  & $\checkmark$  &  82.49\%\\
				&$\checkmark$   & $\checkmark$  & $\checkmark$  &\textbf{85.69\%} \\
				\hline
			\end{tabular}
		}
	}
\end{table}

\begin{figure*}[!t]
	\centering \includegraphics[width=6.7in]{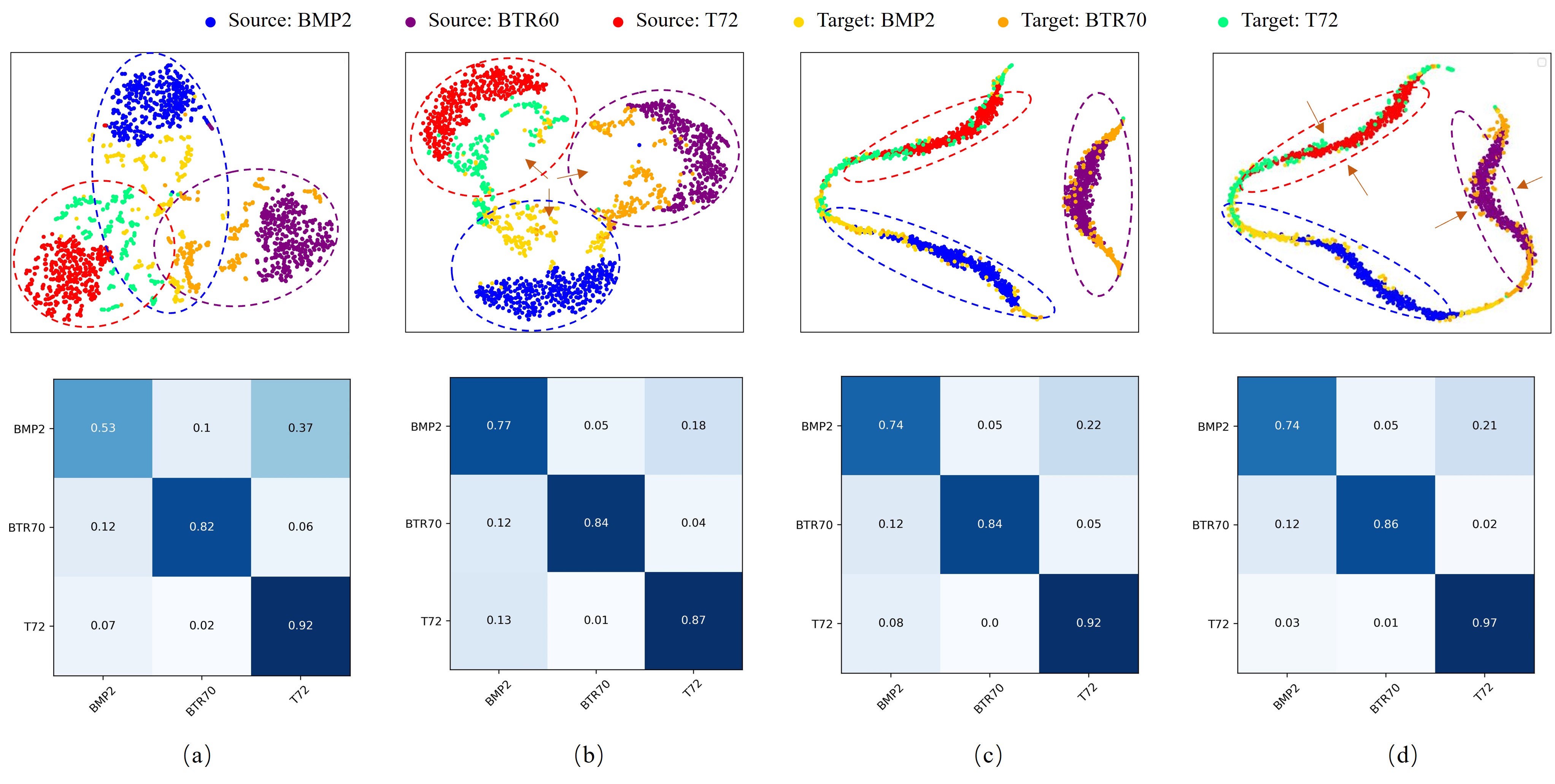}%
	\caption{Visualisation of t-SNE in SD and TD features after adding distinct modules. 10 randomly selected samples from each class in the measured-train. (a) Initial version without any modules. (b) With the addition of the TAIS module. (c) Add TGKT to (b). (d) Integration of all modules (TAIS, TGKT and CRKT). Their recognition accuracies are 75.64\%, 82.64\%, 83.13\%, and 85.69\%, respectively:}
	\label{figE3}
\end{figure*}

\subsubsection{Ablation experiments among various modules}
Table \ref{t3} illustrates the performance gain after adding different modules.
Obviously, the network achieves 77.78\% recognition results by training with only limited measured samples.
If the FSCM data is imported directly to fine-tune the network, the network performance is significantly reduced by 75.54\%.
This phenomenon proves that there exists the distribution gap between the domains, unnegligiblly.
To verify the validity of each module loss, we inserted them into the network as independent unit, respectively.
The incorporation of the TAIS module brings the recognition accuracy to 82.62\% and performance gains of about 5\%.
That is purely indebted to, the filtering of redundant information by the TAIS module, and the implicit elimination of the inter-domain gaps.
Then, only the TGKT module is inserted in the network and the overall accuracy reaches 81.37\%.
That means, with the help of TGKT, the invariant features in the cross domains can be extracted to supplement target generic knowledge.
Finally, the CRKT module migrates the category relation information from the SD to the TD, improving the network performance by more than 2\%, reaches 80.41\%.

The above experiments proved that all the proposed modules deliver information gain to the TD from various perspectives.
Given the distribution gaps between various domains, the MDD metric function is adopted as the center to validate the effect of coupling various modules.
After coupling TAIS and TGKT, the recognition performance reaches 83.13\%.
Domain invariant features that strongly associate with the task are selected.
Then, on top of TGKT, the CRKT module is inserted to ensure inter-domain category semantic information, and the recognition accuracy reaches 82.49\%.
Lastly, we involve all modules together in the network to address the full migration of SD knowledge in terms of beneficial feature screening, domain-invariant feature extraction, and category relation knowledge migration.
Ultimately, the overall performance is increased over 7\% to 85.69\%.

\subsubsection{Visualisation analysis of various modules}
Additionally, to demonstrate the impact by adding various modules to the network more intuitively, we perform the T-distributed Stochastic Neighborhood Embedding (t-SNE) \cite{JMLR:v9:vandermaaten08a} visualisation of the features in each stage, as shown in Fig. \ref{figE3}.
The initial experiment is the method without adding any of the proposed modules, as shown in Fig.\ref{figE3}(a).
Clearly, it is easy to distinguish samples from SD or TD, and TD samples suffer from poor separability.
Subsequently, the integration of TAIS module makes the inter-domain feature distributions have mutual proximity.
And, the boundaries between diverse categories of measured samples become sharper.
This phenomenon explained that the TAIS module increases the separability of TD features by augmenting category information, and implicitly closes the inter-domain distance.
Given that the MDD loss is mainly in charge of confusing the inter-domain distribution, it is difficult to accurately determine the origin (SD or TD) of the samples in the same class in Fig. \ref{figE3}(c), and the intra-class density and inter-class separability increase.
However, given the mismatch in the number of samples between domains, it would make MDD focus more on the aggregation of samples from the SD in the optimization phase, which would be prone to overkill.
CRKT constrains this phenomenon by maintaining the consistency of category structure relation between domains.
Compared to the original unadapted method, the proposed method successfully performs the knowledge migration from all categories, especially the BMP2 and the BTR70, as shown in Fig. \ref{figE3}(a) and (d).

Through ablation experiments and visualisation analysis of various modules, all the proposed elements yield network performance gains and achieve the all-round migration of SD knowledge.

\begin{table*}[!t]
	\renewcommand\arraystretch{1.4}
	\caption{Performance comparison of different limited data conditions.\label{tE5}}
	\centering
	\resizebox{\textwidth}{!}{
		\setlength{\tabcolsep}{1.8mm}{\begin{tabular}{c c c c c c c c c c c} \hline
				\multirow{2}{*}{ } &\multirow{2}{*}{Method}  &\multicolumn{9}{c}{Measured Samples Size} \\
				\cline{3-11}
				&  & 1 & 5 & 10 & 20  & 40 & 60 & 100 & 150 & ALL\\
				
				\hline
				\multirow{4}{*}{Target only}   & AconvNet \cite{7460942}  & 44.78\%    & 49.23\%       & 77.78\%     & 80.64\%  & 89.79\% & 95.57\% & 96.76\% & 96.92\% & 100\% \\  
				& ResNet \cite{7780459}     & 50.25\%    & 55.20\%       & 71.38\%     & 79.05\%  & 84.16\% & 87.05\% & 92.67\%\ & 94.38\% & 99.65\%\\
				& VGGNet \cite{simonyan2014very}     & 52.47\%    & 55.87\%       &62.60\%  &73.25\%     & 83.82\%  & 91.99\% & 96.06\% & 97.44\% & 99.49\%\\
				& GoogleNet \cite{7298594}  & \textbf{53.49}\%    & \textbf{56.39}\%       & 66.44\%    & 68.31\%  & 77.68\% & 85.86\% & 92.50\% & 94.21\% & 99.66\% \\
				
				\hline
				\multirow{3}{*}{Traditional DA}       & SHFA \cite{6587717}     & 45.83\%    & 63.59\%       & 75.26\%     & 85.35\%  & \textbf{91.82\%} & 93.27\% & 95.57\% & 95.91\% & 100\%\\
				& CDLS \cite{7780918}     & 47.19\%    & 49.57\%       & 68.65\%     & 80.92\%  & 86.37\% & 90.46\% & 94.72\% & 96.25\% & 98.64\%\\
				& CDSPP \cite{2022108362}  & 44.80\%    & 69.51\%       & 77.80\%     & 80.64\%  & 84.27\% & 92.61\% & 95.47\% & 96.34\% & 100\%\\
				\hline
				\multirow{6}{*}{Deep learning DA}       & MRAN \cite{ZHU2019214}     & 36.63\%    & 38.95\%       & 44.73\%     & 52.23\% & 61.56\% & 74.53\% & 83.26\% & 87.11\% & 92.37\%\\
				& Deep Coral \cite{10.1007/978-3-319-49409-8_35}     & 36.12\%    & 38.96\%       & 41.57\%     & 50.03\% & 56.00\% & 60.94\% & 69.55\%  & 75.21\%  & 87.46\%\\
				& DANN \cite{ajakan2014domain}     & -     & 38.16\%       & 42.83\%     & 46.00\% & 52.43\% & 57.82\%  & 63.39\% & 71.65\% & 89.63\%\\
				& DAN \cite{pmlr-v37-long15}     & 48.89\%    & 50.77\%       & 65.25\%     & 71.72\% & 78.56\% & 85.98\% & 90.18\% & 93.37\% & 98.51\%\\
				& SSAN \cite{Li_2020}     & 47.53\%    & 51.15\%       & 79.39\%     & 86.03\% & 90.65\% & 93.42\% & 95.91\% & 97.27\% & 100\%\\
				\hline
				\multirow{2}{*}{Proposed}       & MHKT     & 50.20\%    & 52.30\%      & \textbf{85.69\%}     & \textbf{ 87.73\%} & {91.63\%} &  \textbf{96.59\%} &  \textbf{97.74\%} &  \textbf{98.24\%} &  \textbf{100\%}\\
				& vs. AconvNet     & $\uparrow{5.42\%}$    & $\uparrow{3.07\%}$      & $\uparrow{7.91\%}$     & $\uparrow{6.09\%}$ & $\uparrow{1.84\%}$ & $\uparrow{1.02\%}$ &  $\uparrow{0.98\%}$ & $\uparrow{ 1.32\%}$ & $\uparrow{0\%}$\\

				\hline
			\end{tabular}
		}
	}
\end{table*}

\subsection{Comparison experiment on Sim2Mstar}

\begin{figure*}[!t]
	\centering
	\subfloat[]{\includegraphics[width=1.5in]{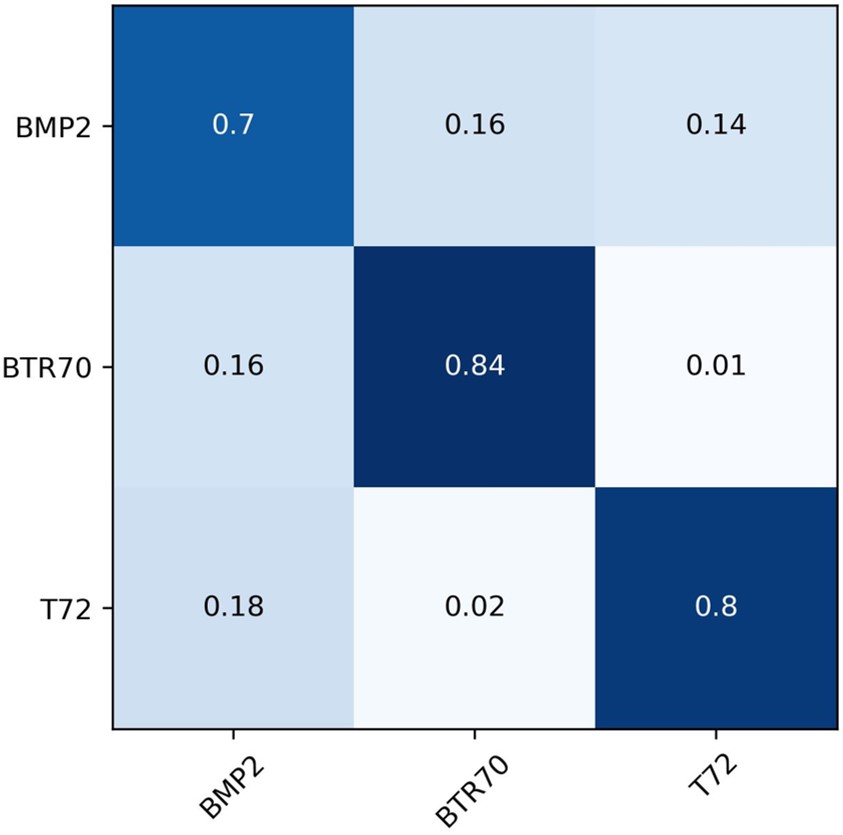}}%
	\label{Real SAR image}
	\hfil
	\subfloat[]{\includegraphics[width=1.5in]{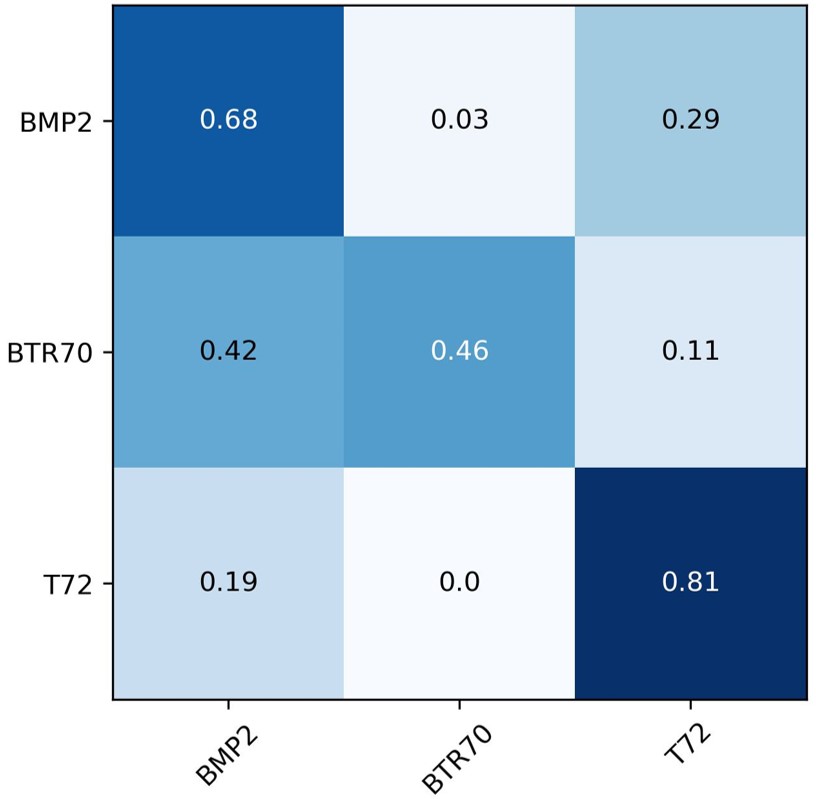}}%
	\label{Real SAR image}
	\hfil
	\subfloat[]{\includegraphics[width=1.5in]{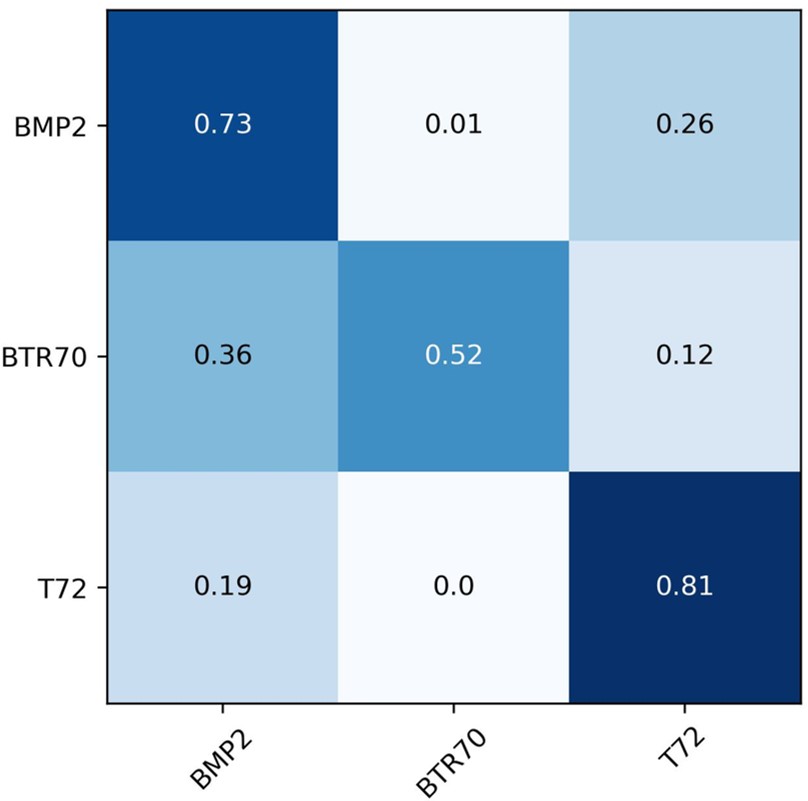}}%
	\label{ASC reconstructed SAR image}
	\hfil
	\subfloat[]{\includegraphics[width=1.5in]{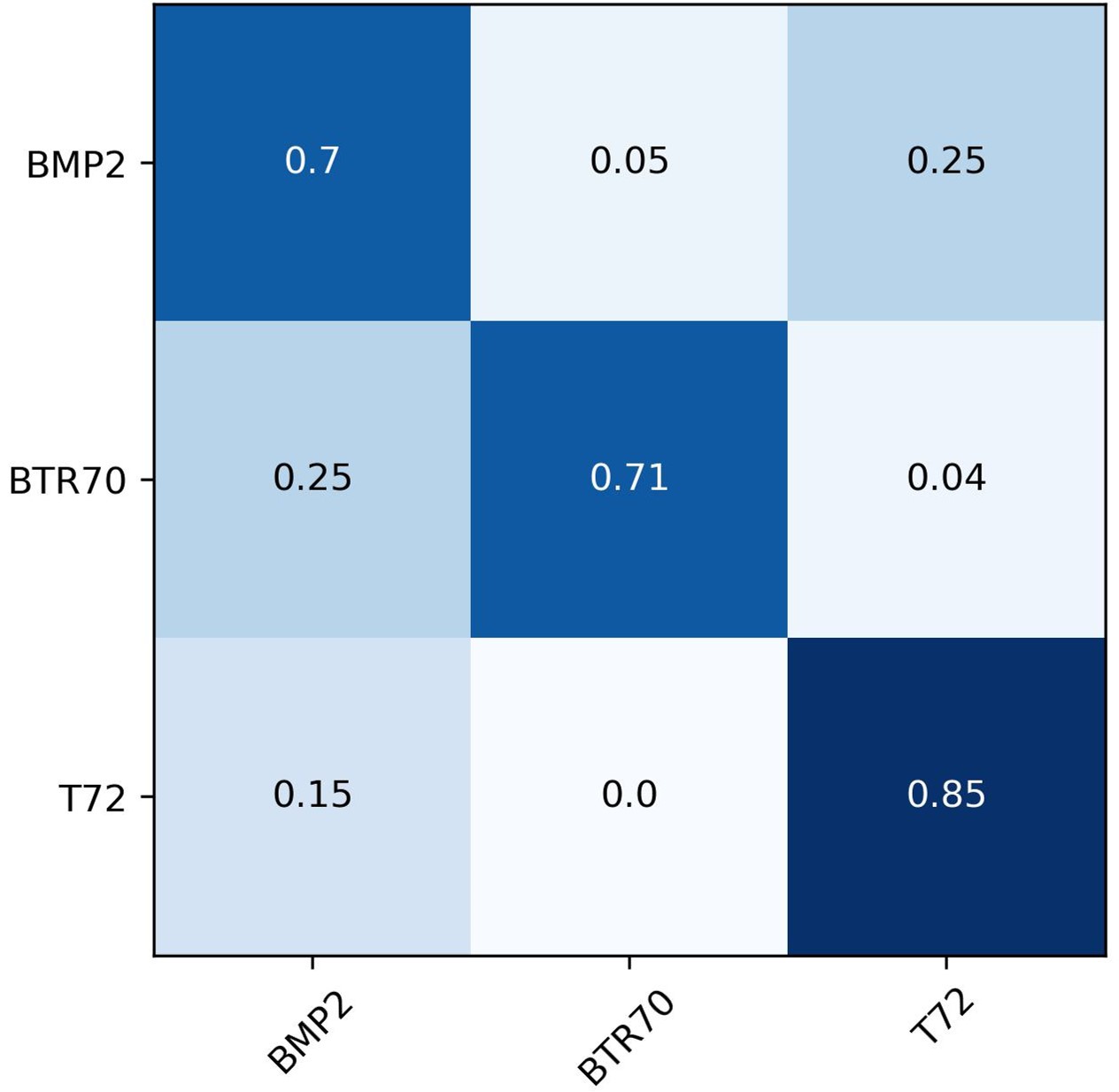}}%
	\label{Real SAR image}
	\hfil
	\subfloat[]{\includegraphics[width=1.5in]{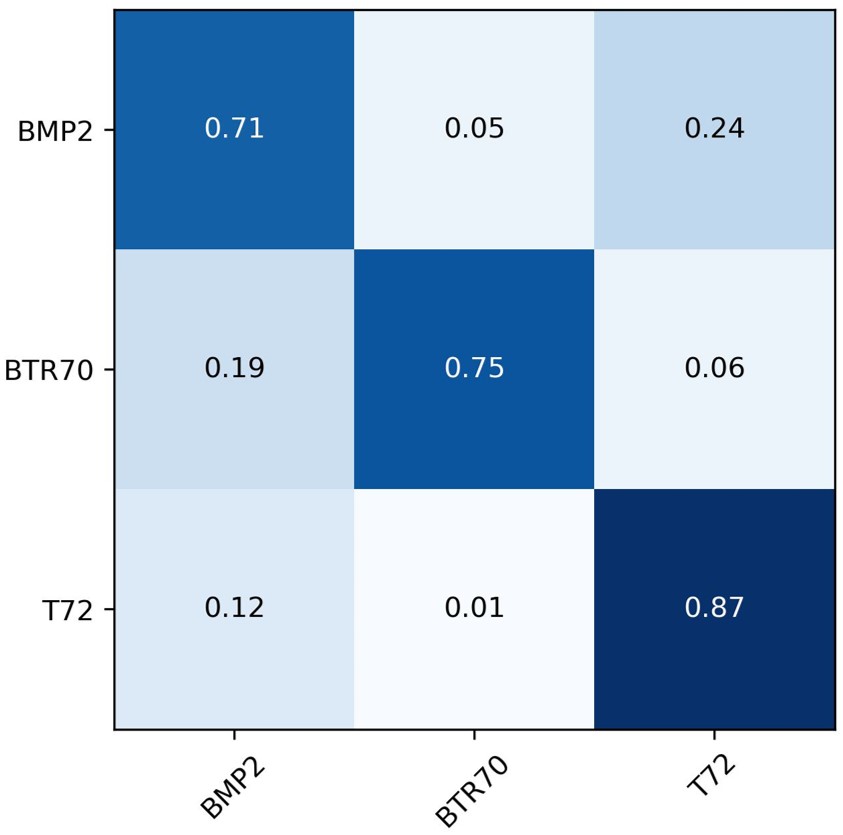}}%
	\label{Real SAR image}
	\hfil
	\subfloat[]{\includegraphics[width=1.5in]{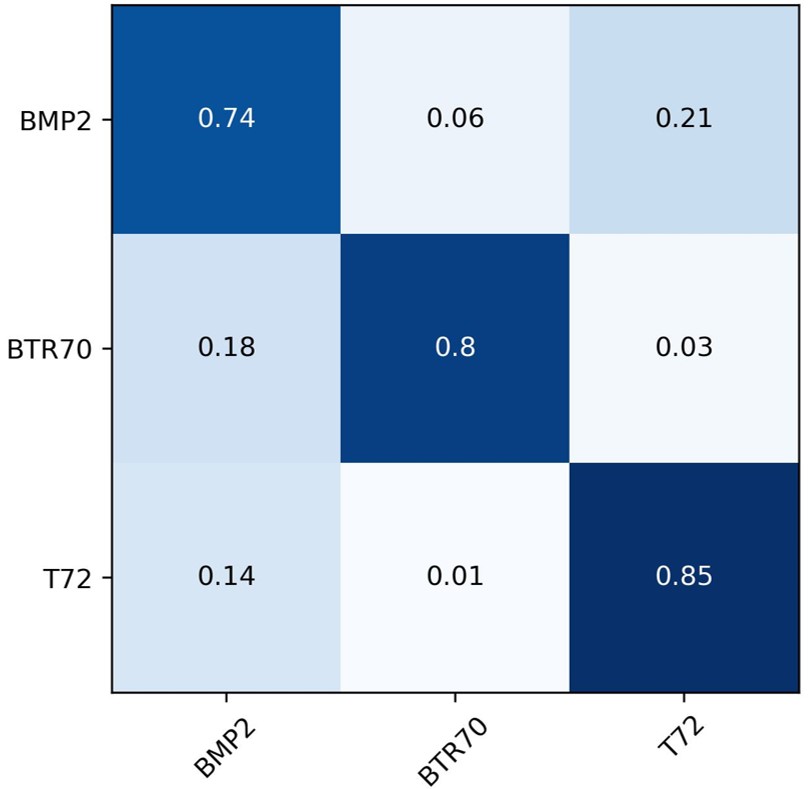}}%
	\label{Real SAR image}
	\hfil
	\subfloat[]{\includegraphics[width=1.5in]{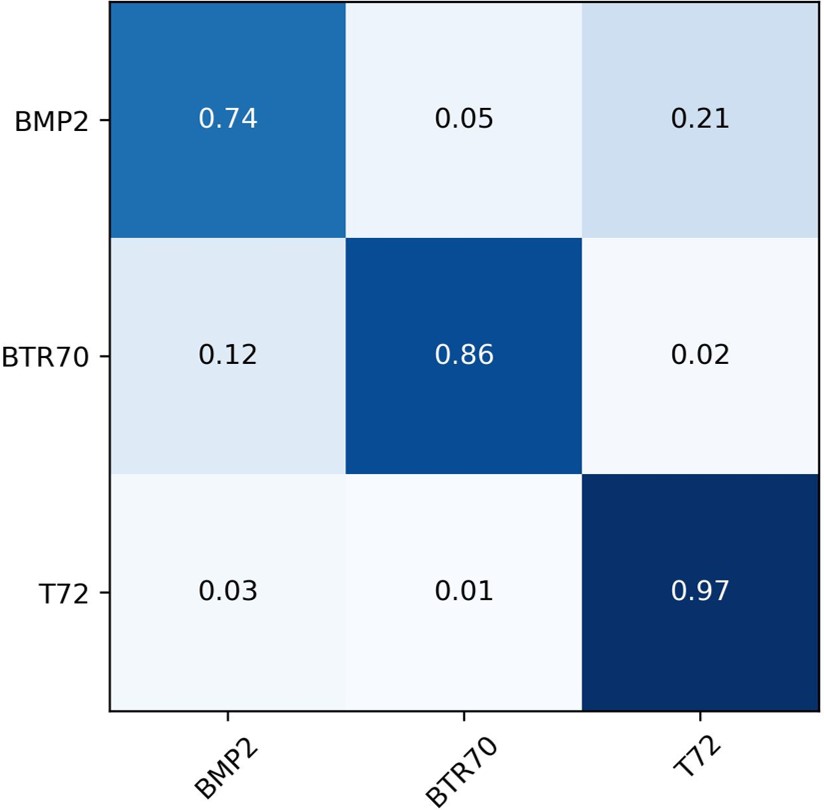}}%
	\label{Real SAR image}
	\hfil
	\subfloat[]{\includegraphics[width=1.3in]{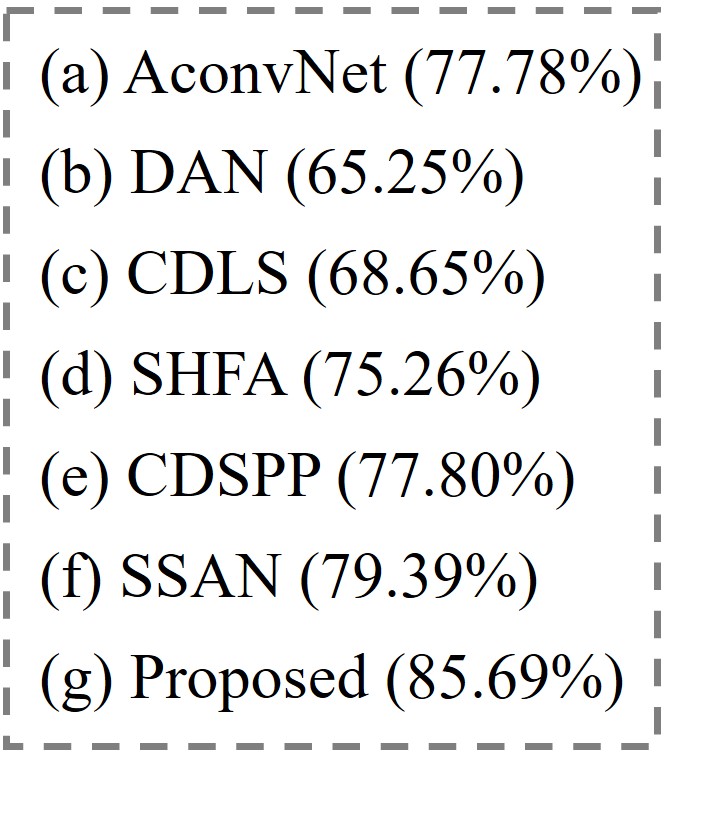}}%
	\label{Real SAR image}
	\caption{Confusion matrix of different methods. 10 randomly selected samples from each class in the measured-train. (a) AconvNet. (b) DAN. (c) CDLS. (d) SHFA. (e) CDSPP. (f) SSAN. (g) Proposed method.}
	\label{figE5}
\end{figure*}

\begin{figure*}[!t]
	\centering
	\subfloat[]{\includegraphics[width=2.3in]{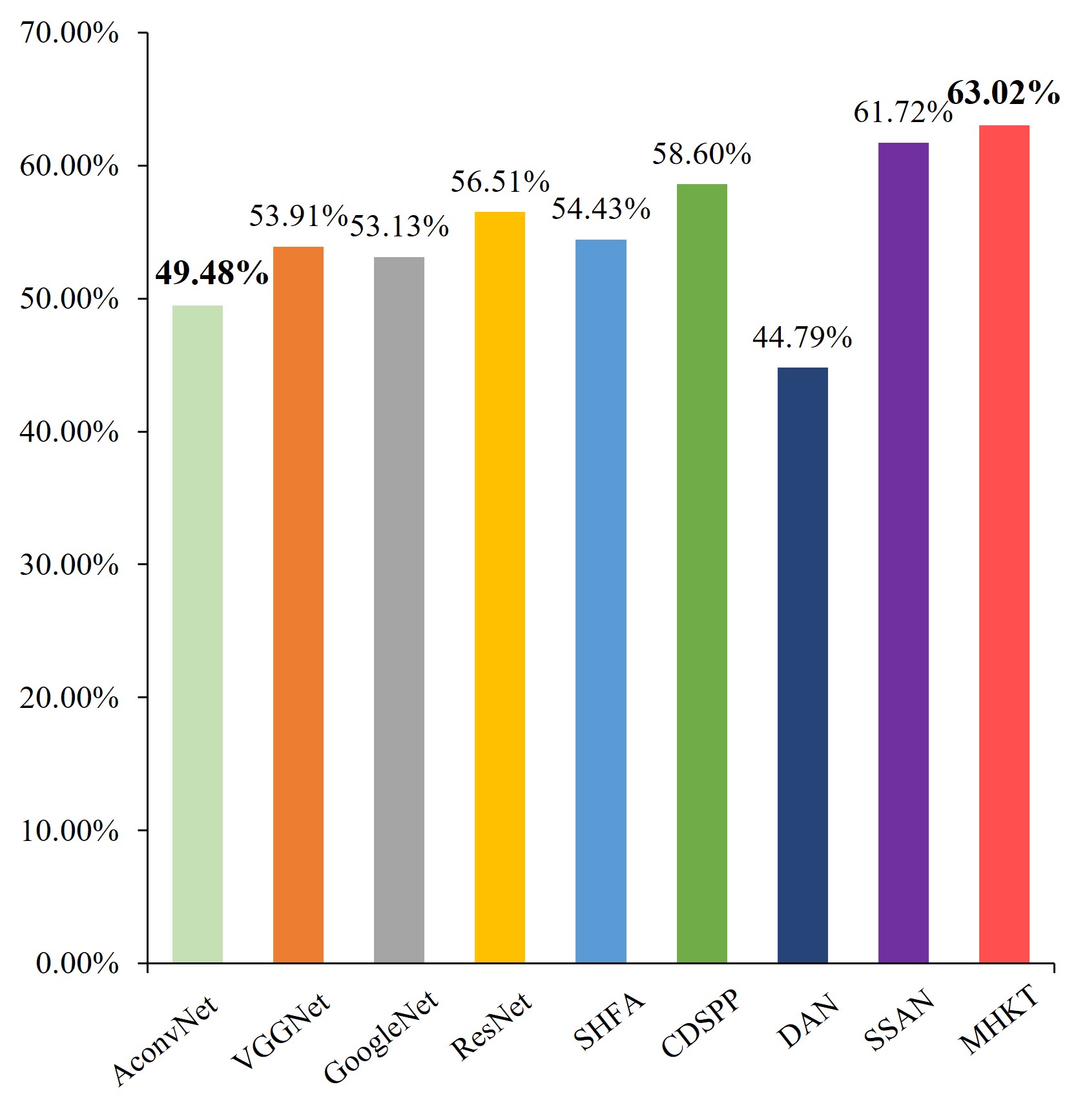}}%
	\label{Real SAR image}
	\hfil
	\subfloat[]{\includegraphics[width=2.3in]{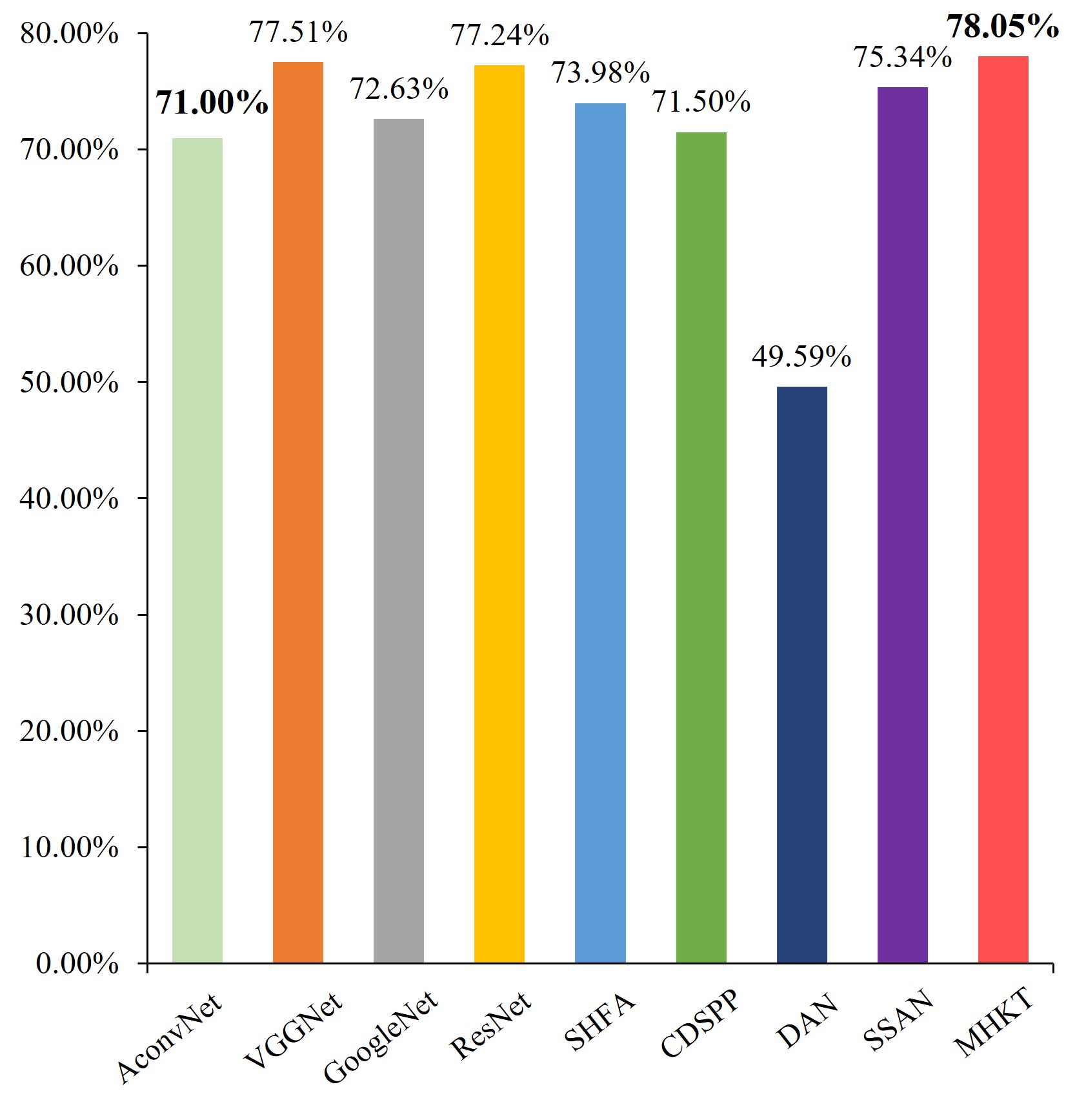}}%
	\label{Real SAR image}
	\hfil
	\subfloat[]{\includegraphics[width=2.3in]{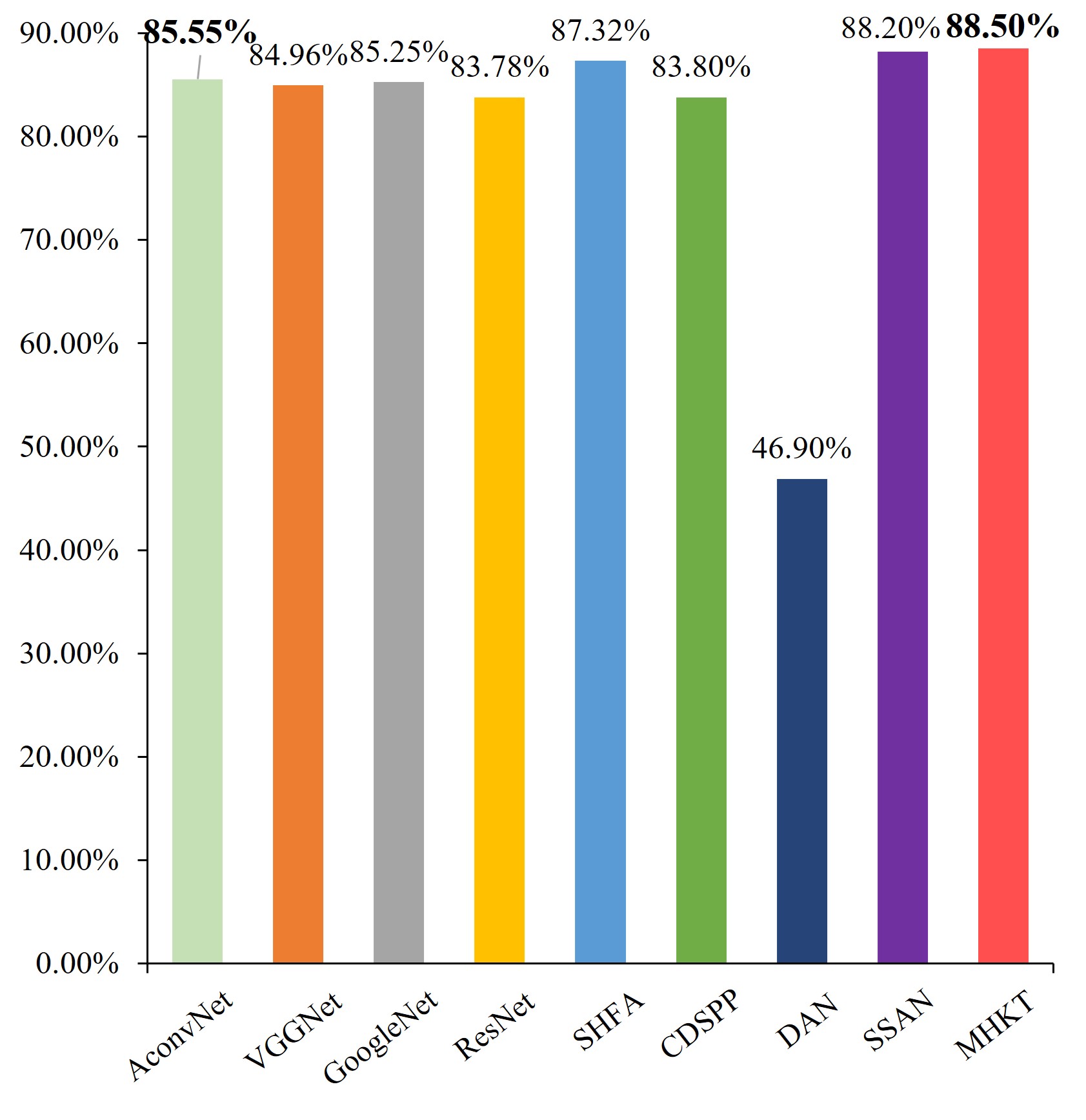}}%
	\label{ASC reconstructed SAR image}
	\hfil
	\subfloat[]{\includegraphics[width=2.3in]{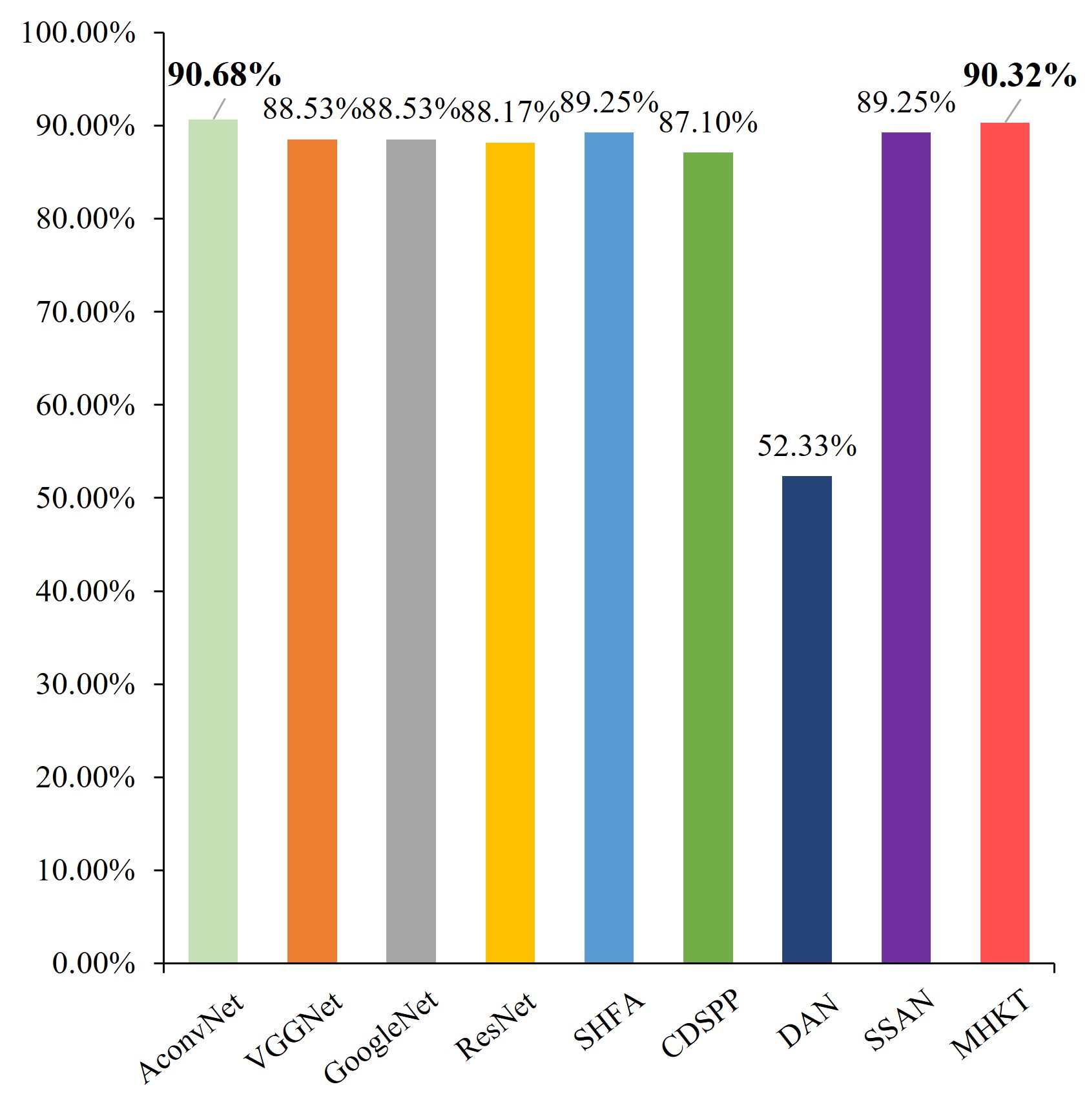}}%
	\label{Real SAR image}
	\hfil
	\subfloat[]{\includegraphics[width=2.3in]{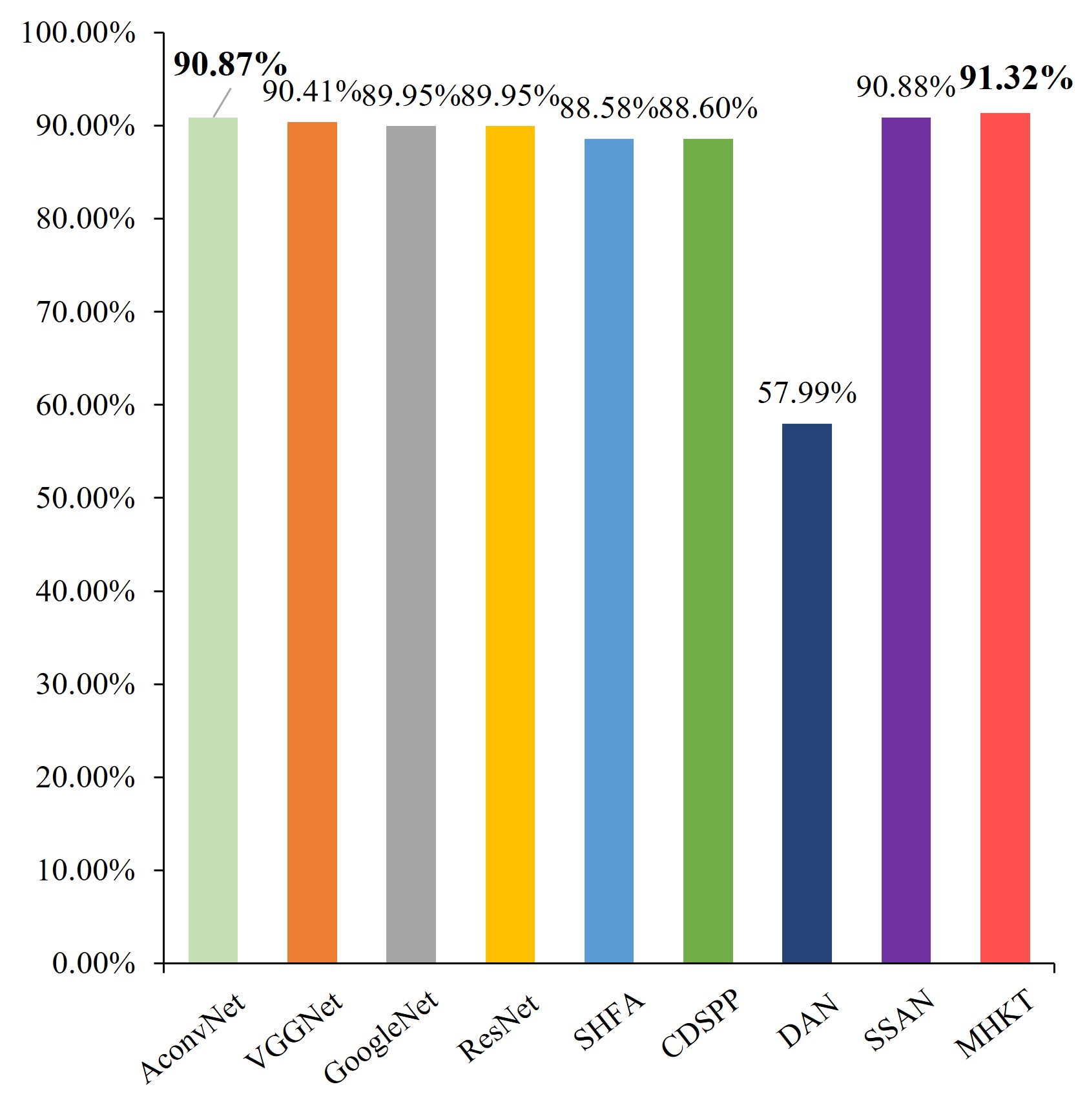}}%
	\label{Real SAR image}
	 \hfil
	\caption{Performance comparison of different limited data conditions. (a), (b), (c), (d), (e) are the experimental results with only 5, 10, 20, 40 and 60 measured SAR aircraft images per class}
	\label{figEF}
\end{figure*}

To further verify the superiority of the proposed method, we compare it with some state-of-the-art methods.
These methods can be divided into three categories: target only, traditional DA, and deep learning DA methods.
Target only methods employ only the limited SAR images as the training set to test the method performance.
Traditional and deep learning DA methods contain a variety of superior domain adaptation methods.
Specifically, we validate the performance of the proposed method in the case of TD with different number of measured samples, as shown in the Table \ref{tE5}.

\subsubsection{Comparative experiments under limited samples}

Firstly, we compare with some typical CNN methods, ResNet, VGGNet and GoogleNet, to verify the fundamental performance of algorithms in the case of limited samples.
Obviously, the general visual network performs better when there are less than 10 samples per class.
As the amount of data increases, the performance of the SAR ATR special network, AconvNet, remains stable and superior.
Particularly, AconvNet completely distinguishes all types of targets when the training data is maximal.
Motivated by this phenomenon, the AconvNet is adopted to extract features from the measured images.
And, the proposed method exhibits performance gains compared to AconvNet in all limited data cases.

Then, we compare with some state-of-the-art traditional HDA methods.
The crucial concept of the traditional methods is to project the SD and TD features into a common feature space.
Specifically, the SHFA method performs relatively steady which ensures effective knowledge migration in most cases.
This is because it preserves the original features on the ground of obtaining common features, which not only enhances the expression of shared features, but also retains more information related to the target.
CDLS and CDSPP map SD features to the TD and common feature space, respectively.
They lack deep processing of original features from various domains and only rely on feature projections to capture domain-invariant features.
Consequently, as the amount of data increases, the ability to discriminate and capture useful features limits their performance, leading to the negative information migration.

Subsequently, we compare with some typical approaches based on deep learning.
Currently, popular deep domain adaptation methods are centred around MMD, Coral distribution metrics and adversarial domain adaptation methods.
MRAN, Deep Coral and DANN are three representative approaches.
Obviously, these methods are ineffective in solving the migration task in this paper.
All of these methods exhibited noticeably negative migration under various training data constraints.
DAN exploits the Multi-kernel MMD metric to measure the distribution gap between domains, which improves its performance compared to the previous three methods, but still fails to outperform the AconvNet method.
SSAN migrates SD category semantic information to TD based on adversarial learning and maintains the consistency of features and category semantics.
It maintains the superior recognition performance generally compared to the traditional methods.
And, this method brings promise for employing deep learning methods to solve the problems.

Markedly, the performance improvement of the proposed method over the AconvNet method is significant.
However, limited by the performance of the AconvNet, the performance of the proposed method is slightly worse than the other target only methods when the sample size is extremely small.
With the increase in the amount of data, the merits of the proposed method gradually emerge.
Notably, our method proves the auxiliary significance of FSCM data for practical recognition tasks under deep learning framework for the first time.

\subsubsection{Confusing matrix analysis of different methods}
Here, we show the confusion matrices of the 6 optimal comparison methods and the proposed methods, as shown in Fig. \ref{figE5}.
We still randomly select 10 samples for each class for training.

Fig. \ref{figE5}(a) illustrates the confusion matrix of AconvNet with a recognition accuracy of 77.78\%.
Evidently, the AconvNet method is more likely to discriminate BTR70 and T72.
Subsequently, Fig. \ref{figE5}(b), (c) and (d) exhibit results from DAN, CDLS and SHFA with recognition rates of 65.25\%, 68.65\% and 75.26\% respectively.
They all perform significantly lower compared to the baseline method, yet there are errors of recognition to various degrees for the BMP2 and BTR70.
The recognition accuracy of CDSPP is 77.80\%, which brings a limited performance gain.
It is first demonstrated that the FSCM is of ancillary significance for SAR ATR, in this paper.
The more advanced SSAN (79.39\%) method shows a significant increase in both BMP2 and T72 recognition accuracy compared to the baseline method, although the BTR70 has a decrease in recognition accuracy, as illustrated in Fig. \ref{figE5}(f).
The above two approaches take a more crude approach for knowledge transfer and thus fail to capture the impact of the transferred knowledge on the different categories.
Notably, the proposed method purifies the task-related beneficial information and designs the distribution alignment loss to aggregate the intra-class features and increase the inter-class differences in a more focused way, achieving the recognition accuracy of 85.69\%.

In summary, the proposed method implements the exploitation of FSCM knowledge to assist the SAR target recognition task in the deep learning framework for the first time.
The experiment results also verify the effectiveness of the proposed strategy.

\subsection{Comparison experiment on Sim2Air}
To verify the generalizability of the proposed method, we perform experiments on the Sim2Air dataset.
Here, we select some methods with better performance for comparison based on the results in Table \ref{tE5}.
We randomly select 5,10,20,40,60 images per class for training, as shown in Fig. \ref{figEF}.
Obviously, the performance gain of the proposed method is more obvious when the number of measured samples is fewer.
Specifically, when only 5 measured samples per class are involved in training, the proposed method achieves the 63.02\% recognition accuracy, which shows the performance gain of 13.54\% compared to the baseline AconvNet (49.48\%).
As the number of measured samples increases, the performances are progressively similar among different algorithms, but the proposed method still maintains the state-of-the-art performance.
This is because, as the amount of data increases, the network may learn more discriminative features which may not necessarily describe the target.
Therefore, the addition of FSCM knowledge provides inconspicuous performance gains.

\section{Conclusion}
In this paper, a novel MHKT network is proposed, which uses FSCM data knowledge to assist SAR images target recognition based on the deep learning framework for the first time.
Firstly, the TAIS module is presented to decouple the task-associated information in the SD and TD, to apply high-level semantic features for adapting and filtering the redundancy information.
Subsequently, in the TGKT module, we propose a novel MDD metric function that aligns the inter-domain marginal and conditional distributions while increasing the intra-class density and enhancing the inter-class discrepancies, resulting in more discriminative domain-invariant features.
Moreover, implicit semantic knowledge is further mined with the aid of the CRKT module, which is helpful to migrate fine-grained category knowledge.
Notably, the MHKT method maintains optimal performance in the two new datasets, especially in the limited samples conditions.

In future work, we will deeply analyse the correlation information between the FSCM data and SAR images primitive features and explore more advanced manners to purify the domain-invariant knowledge.
Moreover, we will produce richer simulated-measured datasets for more adequate research.

\bibliographystyle{IEEEtran}
\bibliography{refs}
\vfill

\begin{IEEEbiography}[{\includegraphics[width=1in,height=1.25in,clip,keepaspectratio]{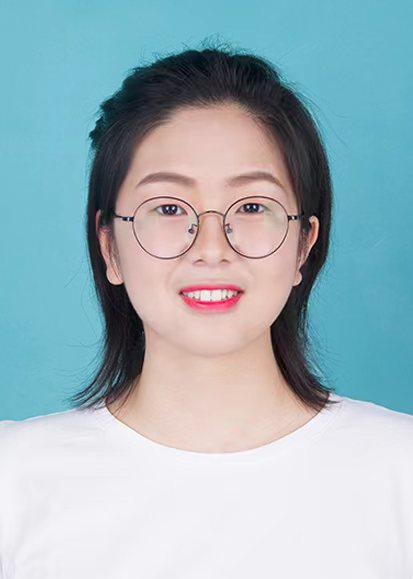}}]{Chenxi Zhao}
	received the B.S. degree and M.S. degree from the Beijing University of Chemical Technology, Beijing, China, in 2018 and 2021, respectively. She is currently working toward the Ph.D. degree in information and communication engineering with the State Key Laboratory of Complex Electromagnetic Environment Effects, National University of Defense Technology, Changsha, China.
	
	Her research interests include SAR image interpretation, feature extraction, and machine learning.\end{IEEEbiography}

\begin{IEEEbiography}[{\includegraphics[width=1in,height=1.25in,clip,keepaspectratio]{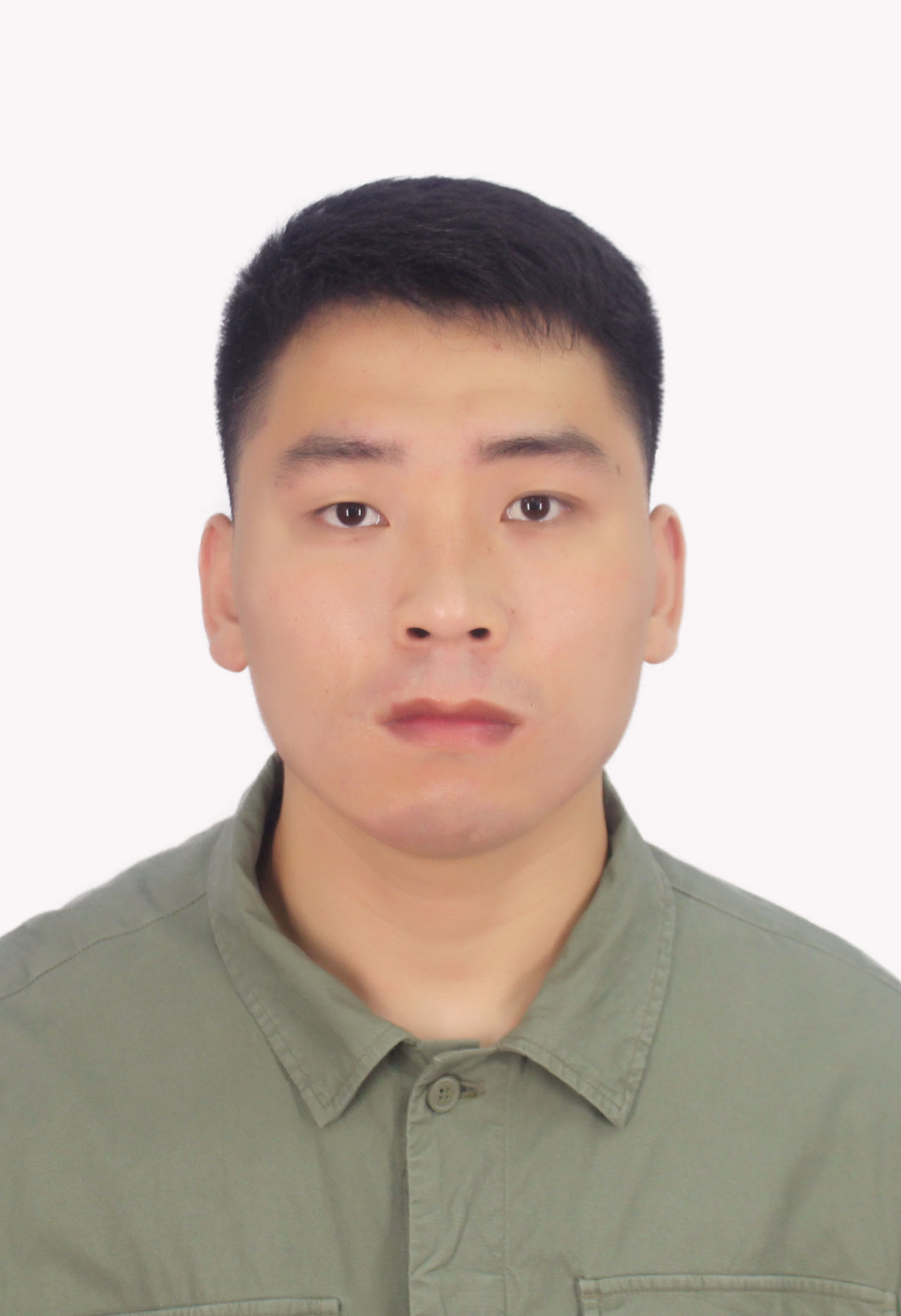}}]{Daochang Wang}
	received the B.S. degree from Qufu Normal University, Rizhao, China, in 2020, the M.S. degree from Beijing University of Chemical Technology, Beijing, China, in 2023, and is currently pursuing a Ph.D. degree at the University of National Defense Technology, Changsha, China.
	
	His research interests include synthetic aperture radar image processing, target detection, and human action recognition.\end{IEEEbiography}

\begin{IEEEbiography}[{\includegraphics[width=1in,height=1.25in,clip,keepaspectratio]{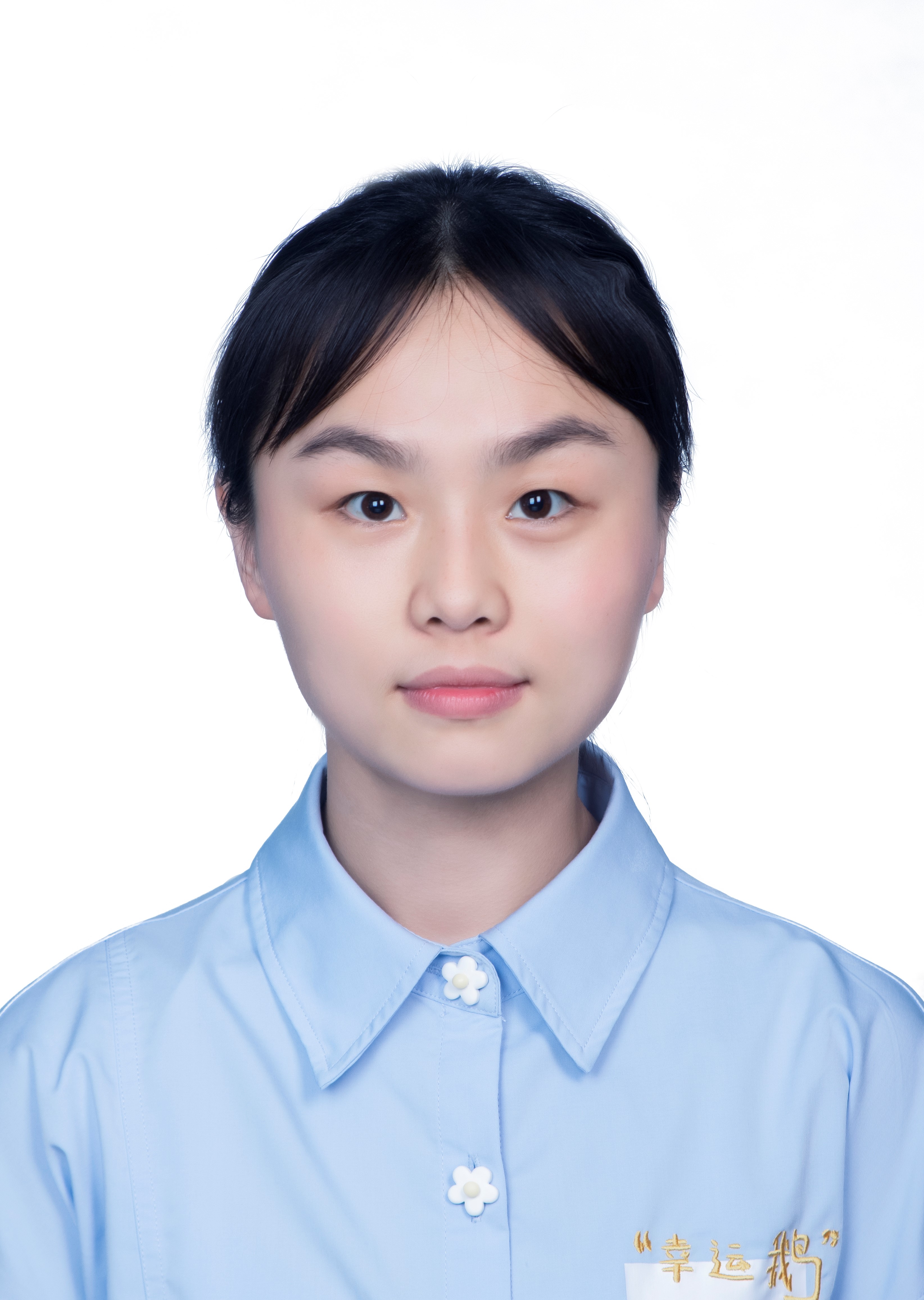}}]{Wancong Li}
	was born in Henan, China, in 1999. She received the B.S. degree in communication engineering from Central China Normal University, Wuhan, China in 2021. She is currently working toward the Ph.D. degree at the School of Electronic Information, Wuhan University, Wuhan, China. Her research interest includes electromagnetic theory and its application, radar imaging, and complex objects characterizing.\end{IEEEbiography}

\begin{IEEEbiography}[{\includegraphics[width=1in,height=1.25in,clip,keepaspectratio]{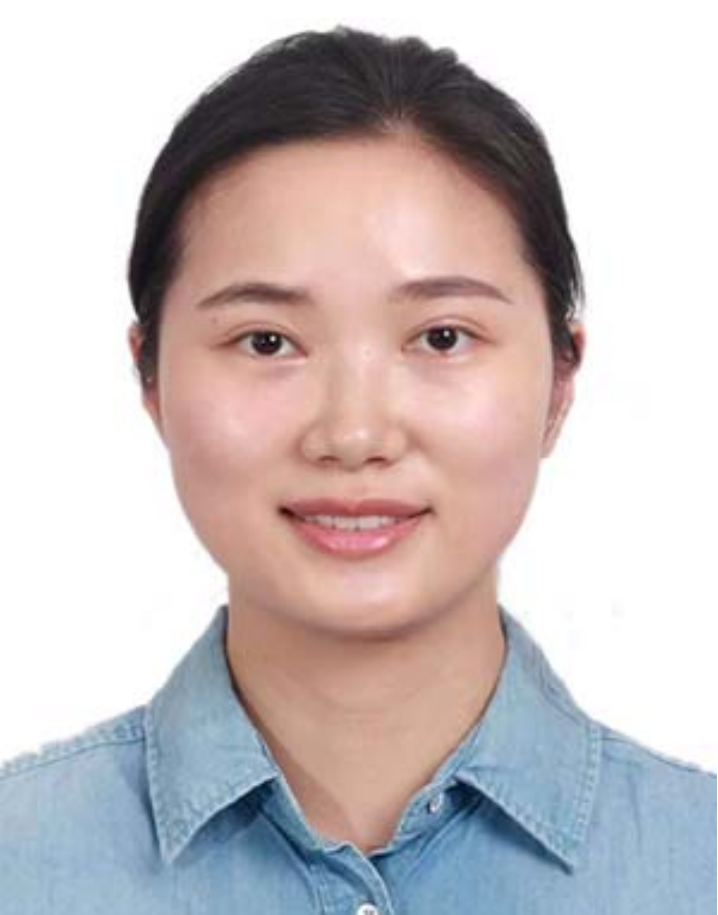}}]{Siqian Zhang}
	received the B.S. degree in electronic information engineering from Hubei University, Wuhan, China, in 2009. She received the M.S. and the Ph.D. degrees from the National University of Defense Technology, Changsha, China, in 2011 and 2015, respectively. 
	She is an Associate Professor with the College of Electronic Science and Technology, National University of Defense Technology. 
	
	Her research interests include radar signal processing, SAR imaging, SAR image interpretation and deep learning.
\end{IEEEbiography}

\begin{IEEEbiography}[{\includegraphics[width=1in,height=1.25in,clip,keepaspectratio]{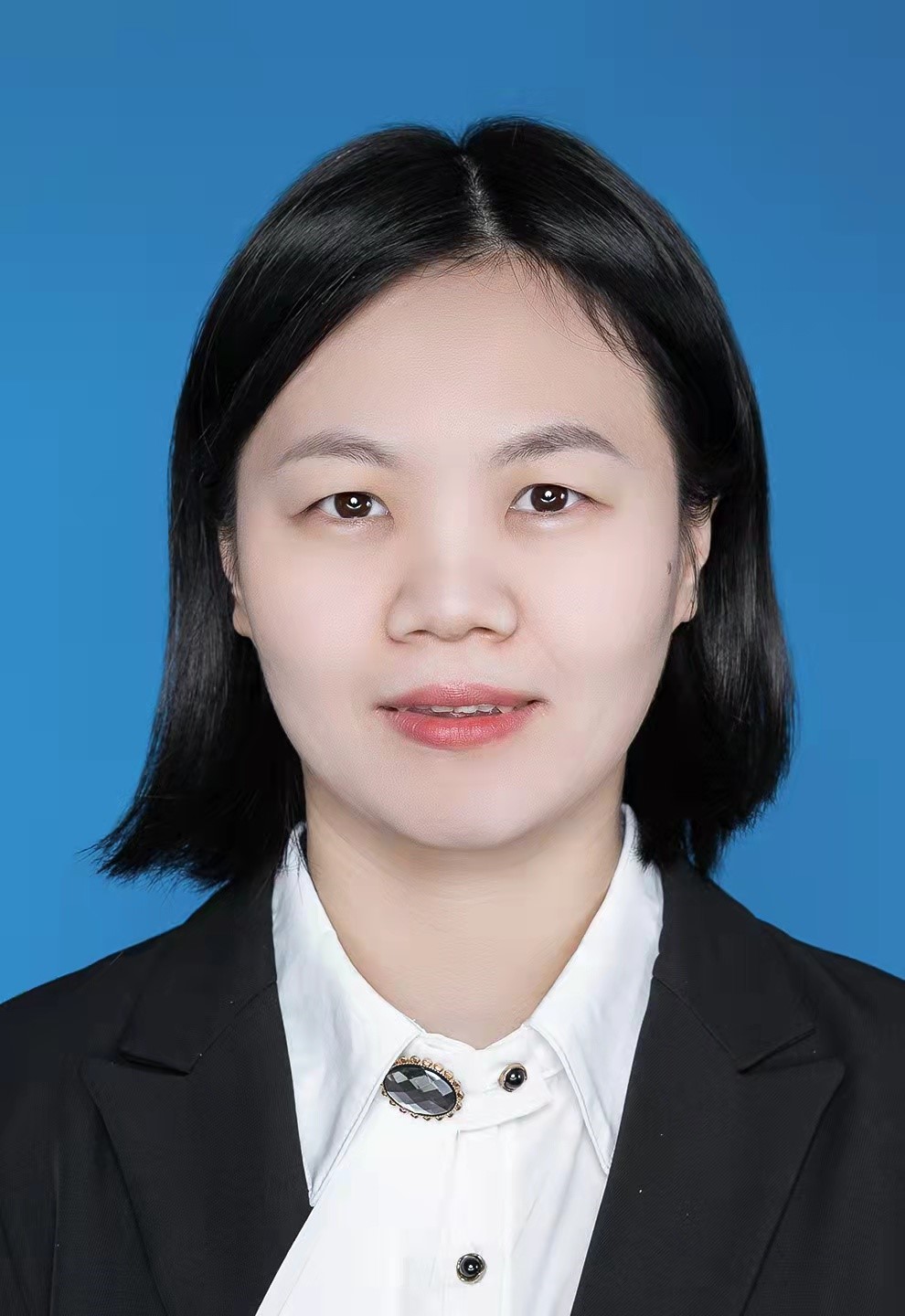}}]{Siyuan He}
	was born in Zhejiang, China, in 1982. She received the B.S. degree in communication engineering and the Ph.D. degree in radio physics from Wuhan University, Wuhan, China, in 2003 and 2009, respectively. From 2005 to 2006, she was a Research Assistant in the Wireless Communications Research Center, City University of Hong Kong, Hong Kong. From 2009 to 2011, she worked as a Postdoctoral Researcher at Wuhan University. She is currently a Professor with the School of Electronic Information, Wuhan University. Her research interest includes electromagnetic theory and its application, computational electromagnetism, radar imaging, and electromagnetic inverse scattering. 
	
	His research interests include machine learning, remote sensing image processing, and multitemporal image analysis.
\end{IEEEbiography}

\begin{IEEEbiography}[{\includegraphics[width=1in,height=1.25in,clip,keepaspectratio]{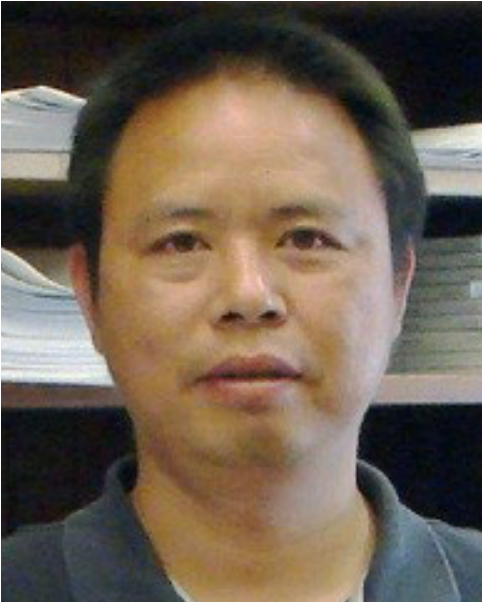}}]{Gangyao Kuang}
	(Senior Member, IEEE) received the B.S. and M.S. degrees in geophysics from the Central South University of Technology, Changsha, China, in 1988 and 1991, respectively, and the
	Ph.D. degree in communication and information from the National University of Defense Technology,
	Changsha, in 1995. He is currently a Professor at the School of Electronic Science, National University of Defense Technology. 
	
	His research interests include remote sensing, SAR image processing, change detection, SAR ground moving target indication, and classification with polarimetric SAR images.
	
\end{IEEEbiography}

\end{document}